\documentclass[acmlarge]{acmart}
\makeatletter
\newcommand{\confshort}{\acmConference@shortname}
\newcommand{\conffull}{\acmConference@name}
\newcommand{\confdate}{\acmConference@date}
\newcommand{\confloc}{\acmConference@venue}
\AtBeginDocument{
  \fancypagestyle{firstpagestyle}{
    \fancyhead{}%
    \fancyfoot[C]{}%
  }
  \fancyhf{}
  \fancyhead[LO]{\@headfootfont\shorttitle}%
  \fancyhead[RE]{\@headfootfont\@shortauthors}%
  \fancyhead[LE]{\@headfootfont\footnotesize \confshort, \confdate, \confloc}%
  \fancyhead[RO]{\@headfootfont\footnotesize \confshort, \confdate, \confloc}%
  \fancyfoot[C]{}%
}
\makeatother
\acmBooktitle{\conffull\@ (\confshort), \confdate, \confloc}

\AtBeginDocument{%
  }

\usepackage{dsfont}
\DeclareMathOperator{\Indicator}{\mathds{1}}
\DeclareMathOperator{\R}{\mathbb{R}}

\usepackage[l3]{csvsimple}

\usepackage{amsmath}

\usepackage{amssymb}
\usepackage{tcolorbox}
\usepackage{enumitem}
\usepackage{soul}

\newcommand{\coloneqq}{\mathrel{\mathop:}=}

\copyrightyear{2026}
\acmYear{2026}
\setcopyright{cc}
\setcctype{by}
\acmConference[FAccT '26]{The 2026 ACM Conference on Fairness, Accountability, and Transparency}{June 25--28, 2026}{Montreal, QC, Canada}
\acmBooktitle{The 2026 ACM Conference on Fairness, Accountability, and Transparency (FAccT '26), June 25--28, 2026, Montreal, QC, Canada}
\acmDOI{10.1145/3805689.3812252}
\acmISBN{979-8-4007-2596-8/2026/06}

\begin{document}

\title{Using predictive multiplicity to measure individual performance within the AI Act}

\author{Karolin Frohnapfel}
\authornote{Both authors contributed equally to this work.}
\affiliation{%
  \institution{University of Tübingen and Tübingen AI Center}
  \country{Germany}
}

\author{Mara Seyfert}
\authornotemark[1]
\affiliation{%
  \institution{University of Tübingen and CZS Institute for Artificial Intelligence and Law}
  \country{Germany}
}

\author{Sebastian Bordt}
\affiliation{%
  \institution{University of Tübingen and Tübingen AI Center}
  \country{Germany}
}

\author{Ulrike von Luxburg}
\affiliation{%
  \institution{University of Tübingen, Tübingen AI Center and CZS Institute for Artificial Intelligence and Law}
  \country{Germany}
}

\author{Kristof Meding}
\affiliation{%
  \institution{University of Tübingen and CZS Institute for Artificial Intelligence and Law}
  \country{Germany}
}

\renewcommand{\shortauthors}{K. Frohnapfel, M. Seyfert, S. Bordt, U. von Luxburg, and K. Meding}

\authorsaddresses{
Authors' Contact Information: Karolin Frohnapfel, karolin.frohnapfel@uni-tuebingen.de, and Mara Seyfert, mara.seyfert@uni-tuebingen.de.
}

\begin{abstract}
When building AI systems for decision support, one often encounters the phenomenon of predictive multiplicity: a single best model does not exist; instead, one can construct many models with similar overall accuracy that differ in their predictions for individual cases. Especially when decisions have a direct impact on humans, this can be highly unsatisfactory. For a person subject to high disagreement between models, one could as well have chosen a different model of similar overall accuracy that would have decided the person's case differently. We argue that this arbitrariness conflicts with the EU AI Act, which requires providers of high-risk AI systems to report performance not only at the dataset level but also for specific persons.
The goal of this paper is to put predictive multiplicity in context with the EU AI Act's provisions on accuracy and to subsequently derive concrete suggestions on how to evaluate and report predictive multiplicity in practice. Specifically: 
(1) We introduce the AI Act's accuracy provisions and argue that incorporating information about predictive multiplicity could serve compliance with specific provisions for providers.
(2) Based on this legally rigorous analysis, we suggest individual conflict ratios and $\delta$-ambiguity as tools to quantify the disagreement between models on individual cases and to help detect individuals subject to conflicting predictions.
(3) Based on computational insights, we derive easy-to-implement rules on how model providers could evaluate predictive multiplicity in practice. 
(4) Ultimately, we suggest that information about predictive multiplicity should be made available to deployers under the AI Act, enabling them to judge whether system outputs for specific individuals are reliable enough for their use case.
By following our suggestions and considering potential disagreement between models, one can ensure that AI-supported decisions are well grounded, thereby fostering greater trust in AI systems used for decision support.
\end{abstract} 

\ccsdesc[500]{Social and professional topics~Governmental regulations}
\ccsdesc[500]{Computing methodologies~Machine learning}
\ccsdesc[500]{Computing methodologies~Philosophical/theoretical foundations of artificial intelligence}
\ccsdesc[500]{Applied computing~Law}

\keywords{Predictive Multiplicity, Model Multiplicity, AI Act, Accuracy, Individual Performance, Compliance, Dataset Multiplicity}

\maketitle

\section{Introduction}\label{sec:introduction}
Modern machine learning systems significantly impact people's lives. Automated systems determine whether someone is admitted to a graduate program, gets invited to a job interview, or receives a particular medical treatment \citep{uk_gov_review, allen2024transplant}. While providers of such systems strive to build systems that 
achieve a high {\em overall} performance --- usually in terms of statistical accuracy --- the individuals whose cases are being decided are concerned about receiving trustworthy, transparent, and accurate decisions in their {\em specific} cases. 
However, while high statistical accuracy of an AI system indicates generally reliable predictions, this does not necessarily translate to the individual level. The reason is the phenomenon of {\it predictive multiplicity} \citep{Breiman_2001_StatisticalModeling, Marx_2020_PredictiveMultiplicity, Watson-Daniels_2023, Semenova_2022_OnTheExistenceOfSimpler, Rudin_2024_AmazingThingsComeFrom}: 
For many machine learning tasks, there is no single best model. Instead, many distinct models achieve similarly high statistical accuracy, yet differ in their predictions for individuals.
 
{\bf In real-world applications, predictive multiplicity is ubiquitous} \citep{Rudin_2024_AmazingThingsComeFrom}. Indeed, as long as a machine learning model makes an error, this error can, in principle, be distributed differently, leading to different models that make incorrect predictions for different individuals \citep{Laufer_2025_WhatConstitutes, Black_2024_TheLegalDuty, Marx_2020_PredictiveMultiplicity}. In practice, different models are obtained simply because of the multitude of decisions in the model development process \citep{Suresh_2021_AFrameworkForUnderstanding}.
Concretely, model providers must select and preprocess data, choose a model class, and determine the objective(s) to optimise, among others. All these choices influence the final model, leading to a large set of potential models, many of which may be similar in terms of statistical accuracy \citep{Black_2024_TheLegalDuty, Black_2022_ModelMultiplicity}. 
As such, an important insight from the literature is that predictive multiplicity is not a failure of the AI system provider but is inherent to the training process of machine learning models and the fact that perfect statistical accuracy cannot be achieved in most cases \citep{Marx_2020_PredictiveMultiplicity, Black_2024_TheLegalDuty, Laufer_2025_WhatConstitutes}. 
Nevertheless, for persons whose case exhibits a high disagreement between models, this situation is unsatisfactory: 
If, under some model, their job application was rejected, it may be that another model with similar statistical accuracy would have approved their case.
In situations where AI-based decisions have substantial consequences for individual persons --- for example, in high-stakes applications affecting human welfare or personal rights --- such arbitrariness is highly problematic.

\begin{figure}
    \centering
    \includegraphics[width=\textwidth]{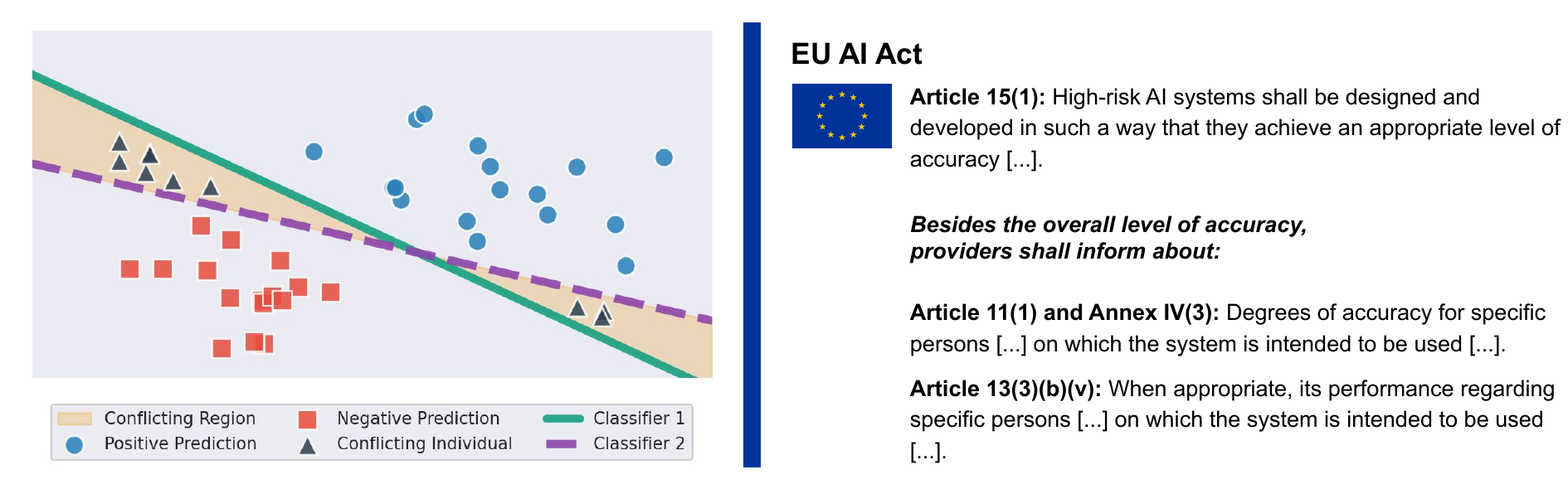}
        \vspace{-0.5cm}
    \caption{\textbf{Our work in a nutshell. }
     Predictive multiplicity means that classifiers of comparable statistical accuracy may decide a large number of individual cases differently (left side). We argue that this challenges the AI Act requirements on accuracy and transparency (right side). To resolve this issue, we recommend that providers and deployers use methods to identify conflicting cases and involve human oversight in the decision-making process.}
    \label{fig:ModelMultiplicity_AI_Act}
\end{figure}

{\bf In this work, we argue that predictive multiplicity raises significant {regulatory concerns}.} We examine predictive multiplicity through the lens of the European Union's AI Act (AIA, Regulation (EU) 2024/1689). The AI Act is a comprehensive piece of legislation that regulates AI within the European market. Particularly relevant to us, the AI Act imposes {\it accuracy obligations on providers of high-risk AI systems.} Specifically, it requires high-risk AI systems to be ``designed and developed in such a way that they achieve an appropriate level of accuracy, robustness, and cybersecurity, and that they perform consistently in those respects throughout their lifecycle'' (Art.~15(1) AIA).
Translating this provision into measurable objectives for the model development process poses significant challenges \citep{nolte2025robustness}, 
and the development of harmonised standards is still ongoing \citep{EU_2023_StandardisationRequest, EU_2025_StandardisationRequest, JTC21_2025_WorkProgramme, FLI_2025_StandardSettingOverview, Hacker_2025_SimplifyingStudy}. 
Nevertheless, we argue that important conclusions can already be drawn about for \textit{whom} accuracy should be assessed under the AI Act. 
In particular, the AI Act requires providers to report accuracy not only on the overall level of entire datasets but also for specific individuals (Art.~11(1) and Art.~13(3)(b)(v) AIA; Annex IV(3)).
If different models yield highly conflicting predictions for an individual, these models are arguably incapable of making a reliable prediction for that individual --- despite their high statistical accuracy. 
We therefore consider predictive multiplicity an important aspect of individual performance.

{\bf Predictive multiplicity provides an appropriate technical framework to measure individual performance.}
In our paper we propose to use the amount of conflict related to an individual as a notion of individual performance. 
The benefit of predictive multiplicity over alternative technical approaches, for example uncertainty quantification, is that predictive multiplicity can be seamlessly introduced into any modelling pipeline \citep{DAmour_2022_UnderspecificationPresents}. It only requires to retrain the model, 
without any further assumptions
or restrictions. 
This is particularly important in the context of broad regulations like the AI Act, which apply to many different machine learning setups. 
An analysis of predictive multiplicity may even be feasible during audits.
The technical challenge of predictive multiplicity is to identify the set of models with comparable overall performance. However, we show below that a simple and computationally feasible ad-hoc approach can already suffice. 
Supplement~\ref{app:why_multiplicity} discusses the technical rationale behind predictive multiplicity in more detail and offers a comparison to uncertainty quantification methods \citep{hullermeier2021aleatoric}.

\vspace{0.2cm}
{\bf Our contributions are as follows: }
\vspace{-0.1cm}
\begin{itemize}
    \item {\bf Legal analysis: }We analyse the AI Act's accuracy and transparency provisions for high-risk AI systems and show that it requires performance evaluations not only on the level of entire datasets, but also on the level of specific individuals (Section~\ref{sec:AIAct}).
    \item {\bf Technical analysis: }We introduce two metrics and a pragmatic computational framework to assess conflicts both at the individual and dataset level (Section~\ref{sec:technical_section}). 
    \item {\bf Overall recommendation: }Ultimately, we recommend providers of high-risk AI systems to deliberately search for models of comparable statistical accuracy --- or other relevant notions of overall performance --- and to make information about conflicting predictions accessible to deployers (Section~\ref{sec:Discussion}).
\end{itemize}

\section{Related work}
\paragraph{\bf Predictive multiplicity}
\citet{Breiman_2001_StatisticalModeling} was among the first to describe the Rashomon effect, which eventually became known as predictive multiplicity \citep{Marx_2020_PredictiveMultiplicity}. Since then, many have explored the phenomenon. We refer to \citet{Ganesh_2025_TheCuriousCase} for a comprehensive review. Most relevant to us are attempts to measure and describe the set of comparably good models \citep{HsuCalmon_2022_RashomonCapacity, Marx_2020_PredictiveMultiplicity, Watson-Daniels_2023, Cooper_2024_ArbitrarinessAndSocialPrediction, Fisher_2019_AllModelsAreWrong, GeorgeNeedellUstun_2025_ObservationalMultiplicity}. Here, global measures like ambiguity
\citep{Marx_2020_PredictiveMultiplicity}
focus on the multiplicity observed for a full dataset, while local measures like the Rashomon capacity \citep{HsuCalmon_2022_RashomonCapacity} or self-consistency \citep{Cooper_2024_ArbitrarinessAndSocialPrediction} quantify the amount of multiplicity per data point. Calculating these measures requires access to the Rashomon set, that is, the set of comparably good models.
Unfortunately, a complete enumeration of the Rashomon set is computationally infeasible for complex model classes. However,
\citet{Xin_2022_ExploringTheWholeRashomonSet} enumerate the Rashomon set for sparse decision trees and \citet{Marx_2020_PredictiveMultiplicity} calculate the ambiguity of the complete Rashomon set of linear classifiers via mixed integer programming.
More complex model classes require approximations of the Rashomon set; existing approaches are usually tailored to specific model classes
\citep{HsuCalmon_2022_RashomonCapacity, Hsu_2024_DropoutBasedRashomonSet, Zhong_2023_ExploringAndInteracting}.
Up until now, the only approach that works for any model class and any dataset is the ad-hoc approach, that is, comparing models that have been trained with different initialisations, hyperparameters, or variations of the training data %
\citep{Marx_2020_PredictiveMultiplicity, Kulynych_2023_ArbitraryDecisions, Cooper_2024_ArbitrarinessAndSocialPrediction}.

Prior work discusses how the non-existence of a single best model creates the challenge of how to reasonably choose from the set of viable options \citep{Black_2022_ModelMultiplicity}.
At the same time, it has been argued that predictive multiplicity allows to optimise for other task-relevant objectives without sacrificing statistical accuracy, such as interpretability, simplicity, or notions of fairness \citep{Semenova_2022_OnTheExistenceOfSimpler, Rudin_2024_AmazingThingsComeFrom, Black_2022_ModelMultiplicity, Black_2024_TheLegalDuty, Laufer_2025_WhatConstitutes}. Yet another strand of research tries to identify the sources of predictive multiplicity
\citep{Cooper_2024_ArbitrarinessAndSocialPrediction, GeorgeNeedellUstun_2025_ObservationalMultiplicity, Meyer_2023_TheDatasetMultiplicityProblem} and investigates how they influence the resulting Rashomon set \citep{Boner_2024_UsingNoise}.

\paragraph{\bf Legal implications of predictive multiplicity}
Previous works have already pointed out how the law should recognise the training of machine learning models as a non-deterministic process: Rather than focusing on single models, sets of possible models and the range of different predictions they might produce for individuals should be considered \citep{cooper2022non, Cooper_2024_ArbitrarinessAndSocialPrediction, Marx_2020_PredictiveMultiplicity}. To measure the consistency between model predictions, \citet{Cooper_2024_ArbitrarinessAndSocialPrediction} introduced self-consistency, a metric similar to our conflict ratio. Our work extends this: By taking into account the accuracy and transparency provisions of the AI Act for high-risk AI systems, we formulate clear recommendations on how information about predictive multiplicity can help achieve compliance with the AI Act.
Other works addressed the question of how to select an appropriate model from a set of comparably good models. Especially in light of negative outcomes, there is an increased need to justify why no other model was chosen that would have yielded a more desirable outcome for the affected individual \citep{Black_2022_ModelMultiplicity}. While variance in outcomes is not necessarily bad but can promote diversity, models must be chosen carefully \citep{Gur_2025_ConsistentlyArbitrary}.
Predictive multiplicity has already been discussed in the context of less discriminatory alternatives, suggesting that firms should explicitly search the Rashomon set for models that do not exhibit indirect discrimination \citep{Black_2024_TheLegalDuty, Laufer_2025_WhatConstitutes}. 

\paragraph{\bf Accuracy in the AI Act} There is little previous work on the topic of accuracy in the AI Act. Most similar to ours in terms of the legal analysis is \citet{nolte2025robustness}. While \citet{nolte2025robustness} discuss the accuracy provision of Article~15 of the AI Act, they mainly focus on the relationship between accuracy and robustness, and do not offer directly implementable recommendations. Additionally, standards relating to the provisions of Article 15 and the general technical verification requirements of the AI Act have been discussed in the literature \citep{kilian2025european, buscemi2025assessing}. However, a deeper analysis of the accuracy requirements and how they could be implemented in practice is currently missing, which is one of the main goals of our paper.

\section{Understanding accuracy within the AI Act}\label{sec:AIAct}
Understanding the relevance of predictive multiplicity for the accuracy provisions of the AI Act requires understanding the scope and structure of the AI Act. Therefore, we start off by giving a short introduction to the AI Act before turning to an in-depth analysis of the accuracy provisions for high-risk AI systems.

\subsection{Introduction to the AI Act}\label{sec:intro_AIA}
The EU AI Act constitutes the world's first comprehensive legal framework specifically dedicated to the regulation of artificial intelligence (see \citet{kaminski2025american} for a comprehensive introduction). 
Its initial foundations lie in the Ethics Guidelines for Trustworthy AI by the \citet{HLEGAI_2019_EthicsGuidelines}, which subsequently were taken up in the European Commission's White Paper on Artificial Intelligence with the aim of creating a corresponding legal framework at the EU level \citep{Commission_2020_WhitePaper}. This ultimately led to the Commission's legislative proposal in 2021 \citep{Commission_2021_AIAProposal}. The AI Act entered into force on 1 August 2024 and has since been implemented in a phased manner: 
While the provisions for prohibited and general-purpose AI models already apply, most of the provisions --- and in particular most of the provisions for high-risk AI systems --- will only do so on 2 August 2026 or 2027 (Art.~113 AIA). 

With the goal of striking a balance between human-centric, trustworthy AI and innovation (Art.~1(1) AIA), the AI Act takes a risk-based approach: Depending on their risk level, different provisions apply to different kinds of AI systems. 
Particular attention is devoted to high-risk AI systems as defined in Article 6 of the AI Act and Annexes I and III. AI systems falling under the category of high-risk include certain products, such as machinery, toys, or medical devices (Annex I), as well as specific use cases, including education, employment, and access to essential public and private services (Annex~III). AI systems listed in Annex III qualify as high-risk only where they pose a significant risk to health, safety, or fundamental rights \citep{nolte2025robustness, kilian2025european}. (For further detail on regulated AI categories beyond high-risk AI systems, including prohibited AI practices and general-purpose AI models, see Supplement~\ref{App:AI_categories}.)

A key distinction in the AI Act is that between AI models and AI systems. According to the (legally non-binding) Recital 97 of the AI Act, an AI system is the combination of an AI model with other components, such as a user interface.
Most of the AI Act's provisions --- including the provisions for high-risk applications that we consider in this paper --- target AI systems. While it is in principle possible that the provisions on high-risk AI systems extend to other parts of the system, in this paper, we discuss them with respect to AI models as they constitute an essential component of an AI system (Rec.~97 AIA). Hence, we will use the term ``model'' straightforwardly in the technical part.

\paragraph{\bf The New Legislative Framework and the Omnibus proposal}
The AI Act's provisions for high-risk AI systems follow the EU's New Legislative Framework for product safety \citep{BlueGuide2022, NLF}, in which the regulator defines essential requirements, while their translation into harmonised standards is delegated to European standardisation bodies and industry experts. Compliance with such standards gives rise to a presumption of conformity (Art.~40(1) AIA) \citep[Art.~40, mn.~1 and 2]{Martini_2024_Art40} \citep{gorywoda2009theneweuropean, kilian2025european}. For high-risk AI systems, it is the European standardisation organisations CEN and CENELEC (European Committee for Standardization and European Committee for Electrotechnical Standardization) that have been tasked with developing such standards \citep{EU_2023_StandardisationRequest, EU_2025_StandardisationRequest, JTC21_2025_WorkProgramme}. Originally planned to be published in 2025, the development of harmonised standards has been delayed and results are now expected to be published no earlier than mid-2026 \citep{FLI_2025_StandardSettingOverview, Hacker_2025_SimplifyingStudy}. This leaves little to no time for implementation before the provisions for high-risk AI systems take effect on 2 August 2026 (Art.~113, AIA) \citep{Hacker_2025_SimplifyingStudy}. 
The European Commission addresses these issues in its proposed Digital Omnibus on AI, which contains various amendments to the AI Act to enhance clarity and to simplify its implementation \citep{Commission_2025_DigitalOmnibusAI}. However, the Omnibus proposal is still progressing through the ordinary legislative procedure.
Accordingly, the precise scope and final provisions of the amendments remain uncertain, and we therefore concentrate on the established norms by the AI Act.

\paragraph{\bf Provider--deployer relationship and intended purpose}
The AI Act acknowledges different roles along the value chain of an AI product: It makes an important distinction between providers (Art.~3(3) AIA) --- entities that develop AI systems --- and deployers (Art.~3(4) AIA) --- entities that use AI systems under their authority. While these two roles are treated as distinct by default, the same entity can occupy both, for instance when a provider also deploys the system. Moreover, if a deployer substantially modifies an AI system, they take over the responsibilities of a provider (Art.~25(1)(b) AIA), thereby replacing the initial provider (Art.~25(2) AIA).
Most of the AI Act's obligations regarding high-risk AI systems target providers --- including Articles 11, 13, and 15, which we examine more closely in this paper (Art.~16 AIA). It is therefore primarily the providers' responsibility to ensure that they bring only safe systems to market. 
Deployers, in turn, must ensure that they only use a high-risk AI system as intended by the provider (Art.~26(1) AIA). In particular, they should only use the AI system with input data that is relevant and sufficiently representative for the intended use case (Art.~26(4) AIA). 
Central here is the concept of intended purpose, which is common to the New Legislative Framework. Providers need to define the intended purpose in the instructions for use that accompany a high-risk AI system (Art.~13 AIA). According to the Blue Guide, providers need to ensure levels of protection that cover the conditions of use which can be reasonably anticipated \citep[Section~2.8]{BlueGuide2022}.
Provider responsibility therefore only extends to reasonably foreseeable applications, not to arbitrary out-of-distribution uses by the deployers.

\subsection{Legal Methodology}\label{sec:legal_methodology}
In our legal analysis of the AI Act's accuracy and transparency provisions for high-risk AI systems, we follow a doctrinal (black-letter) approach, that is, we base our analysis on the legal text as it is written. This involves in particular textual, systematic, and teleological interpretation. While the first interprets the provisions according to their literal meaning, the second broadens the view to the context within which the provision sits, and the third asks what goal the legislator aimed to achieve. To this end, surrounding provisions, the relevant recitals, and, where applicable, \textit{travaux préparatoires} (preparatory and accompanying documents) are taken into consideration \citep{lenaerts2013say, rosler2012interpretation}. Given the novelty of the AI Act and the fact that the provisions for high-risk AI systems are not yet applicable, case law plays no role in this analysis. Furthermore, the autonomy of EU law applies. This means that, in general, national and international law of the Member States plays no role in its interpretation, for instance with respect to the definition of legal terms \citep{Yaroshevskiy_2010_Auslegungsmethoden}.

\subsection{The importance of accuracy for high-risk AI systems}
One fundamental prerequisite for safe and trustworthy AI is the ability of AI systems to reliably make correct predictions. This is especially true in the high-risk areas where errors could cause severe harm --- as they could negatively impact the health, safety, and fundamental rights of affected persons. In order to prevent such harm and to ensure the proper functioning of high-risk systems, the AI Act demands that:

\begin{quotation}
    ``High-risk AI systems shall be designed and developed in such a way that they achieve an appropriate level of accuracy, robustness, and cybersecurity, and that they perform consistently in those respects throughout their lifecycle'' (Art.~15(1) AIA). 
\end{quotation}

However, it does not provide a clear definition of accuracy \citep[Art.~15, mn.~31]{BeckOK_KI-Recht_2025_Art15}. 
Instead, the appropriate level of accuracy depends on the intended purpose of the AI system and the generally acknowledged state of the art (Art.~8(1) AIA; Rec.~74). 
By this, the AI Act recognises that different use cases that come with different levels and kinds of risk might require different facets of accuracy. 
It extends beyond the common machine learning metric of statistical accuracy --- which usually denotes to the proportion of correct predictions on a given dataset --- and refers to accuracy in the plural, as ``metrics'' (Art.~15(3) AIA). 
The specific metrics to be used have yet to be determined. To this end, the European Commission should, ``in cooperation with relevant stakeholders and organisations such as metrology and benchmarking authorities, encourage, as appropriate, the development of benchmarks and measurement methodologies'' (Art.~15(2) AIA). 
In fact, with ISO/IEC TS 4213:2022, a standard for the assessment of classification tasks already exists \citep{ISO/IEC-TS-4213:2022}. As part of the development of harmonised standards, this standard will be replaced by a newer version and complemented by further standards for different kinds of models and applications \citep{standards-mapping}.
In its official standardisation request to CEN-CENELEC \citep{EU_2023_StandardisationRequest, EU_2025_StandardisationRequest, JTC21_2025_WorkProgramme}, the Commission clarifies its expectations regarding accuracy and explicitly asks for standards that go beyond the common machine learning understanding of statistical accuracy: 

\begin{quotation}
    ```Accuracy' shall be understood as referring to the capability of the AI system to perform the task for which it has been designed. This should not be confused with `statistical accuracy', which has a more limited meaning and is one of several possible metrics for evaluating the performance of AI systems'' \citep[Annex~II, p.~7]{EU_2025_StandardisationRequest}.
\end{quotation}

Accuracy in the AI Act therefore needs to be understood in the broader sense of performance, which is defined as ``the ability of an AI system to achieve its intended purpose'' (Art.~3(18) AIA). This clear reference to the intended purpose --- which in turn is defined as ``the use for which an AI system is intended by the provider, including the specific context and conditions of use~[\dots]'' (Art.~3(12) AIA) --- acknowledges the diverse applications of AI systems. Depending on the concrete use case, the understanding of a well performing system might vary: For example, when using an AI system to predict skin cancer, the focus might lie on making as few false negative predictions as possible. When using an AI system to filter job applications, however, special attention will most likely be paid to counteracting disadvantages for protected groups \citep[Art.~15 mn.~32]{BeckOK_KI-Recht_2025_Art15}. 

The development of such standards faces the major challenge of ensuring comparability between systems. At the same time, the standards need to be specific enough to accommodate the diverse contexts in which AI systems can be used. 
The AI Act's additional specific focus on the protection of fundamental rights further complicates matters: While a standard allows for a clear distinction --- compliant or non-compliant --- the protection of fundamental rights usually requires context-specific proportionality assessments \citep{almada2025eu}. 
The standardisation process has indeed faced delays with
the public announcement of the results now expected in the second half of 2026 instead of 2025 \citep{EU_2023_StandardisationRequest, EU_2025_StandardisationRequest, FLI_2025_StandardSettingOverview, Hacker_2025_SimplifyingStudy}. 
But even with the exact standards still missing, the AI Act already provides important guidance as to for \textit{whom} performance shall be evaluated as part of its requirements on the technical documentation (Art.~11 AIA) and the transparency of high-risk AI systems (Art.~13 AIA).

\subsection{The AI Act's emphasis on individual performance}
The provisions regarding accuracy --- or more general performance --- in Article 15 do not stand isolated. They are closely connected to the transparency requirements set out in Articles 11 and 13 of the AI Act. The provider of a high-risk AI system needs to provide information about the achieved levels of performance to national competent authorities (Art.~11(1) AIA; Annex IV) as well as the deployers of the system (Art.~13(3)(b) AIA). This connection between performance and transparency only seems logical because, after all, the legislators' power to enforce certain performance standards hinges on their ability to verify the correctness of the information provided --- which in turn is determined by the level of third-party control over high-risk AI systems \citep[Art.~15, mn.~6]{Martini_2024_Art15}. 
Notably, both articles provide further information with respect to \textit{whom} performance shall be assessed. They specify that information about expected levels of performance might include information on the individual level. Accordingly, the technical report to the national competent authorities should contain:

\begin{quotation}
    ``Detailed information about the monitoring, functioning and control of the AI system, in particular with regard to: its capabilities and limitations in performance, including the \textit{degrees of accuracy for specific persons} [\dots] on which the system is intended to be used and the overall expected level of accuracy~[\dots]'' (Art.~11(1) AIA; Annex IV(3); emphasis added).
\end{quotation} 

In a similar vein, providers should inform deployers about ``the characteristics, capabilities and limitations of performance of the high-risk AI system'' (Art.~13(3)(b) AIA). This includes information about:

\begin{quotation}
    ``The level of accuracy, including its metrics, robustness and cybersecurity referred to in Article 15 against which the high-risk AI system has been tested and validated and which can be expected~[\dots] [and,] when appropriate, its \textit{performance regarding specific persons} [\dots] on which the system is intended to be used'' (Art.~13(3)(b)(ii) and (v) AIA; emphasis added).
\end{quotation}

Taken together, the analysis of Article 15 alongside Articles 11 and 13 of the AI Act highlights two key aspects: First, accuracy must be understood in the broader sense of performance. It therefore needs to be assessed with respect to the intended purpose of the AI system, \textbf{which might include several possible evaluation metrics} depending on specific context and conditions of use. Second, accuracy must be evaluated not only on the level of entire (test) datasets, but can include performance assessments \textbf{with respect to specific 
individuals}. %
However, it remains open how to determine the specific individuals who should be subject to further evaluations. The AI Act provides no explicit guidance or criteria for selecting such individuals. It is furthermore questionable whether the performance assessments for specific individuals by the provider make sense at all: Especially when the provider and deployer are not the same entity, for example, when the deployer purchases the AI system from the provider, it remains unclear as to whether evaluations of specific individuals during development translate to (similar) individuals during deployment. To ensure relevance, we suggest to \textbf{assess performance for all affected individuals by drawing on the concept of predictive multiplicity.}

\subsection{From legal requirements to technical practice: Assessing individual performance with conflict ratios}\label{sec:connection_legal_technical}
Based on the legal analysis in the previous sections, we suggest that providers supply deployers with a tool that enables conflict assessments on all affected individuals, as a possible means for complying with Articles~11, 13, and 15 of the AI Act. Such a tool could also facilitate the provider--deployer relationship by informing deployers about the capabilities and limitations of the high-risk AI system, as required by Article~14 of the AIA.
Equipped with such a tool, deployers could use information about degrees of conflict to decide whether the system performs reliably enough to be applied to a specific individual. We are aware that this suggestion goes beyond the AI Act's current obligations for deployers. We see it, however, as a possible response to the challenge that it might be only the deployer who knows which individuals a system ends up being applied to.

The degree of conflict among model predictions serves as an indicator of the model class' ability to reliably predict specific instances. As an extreme example, consider a data point for which half of the models with comparable statistical accuracy predict the positive class and the other half predict the negative class. While high statistical accuracy may suggest that these models are generally suitable to classify, say, job applications, this aggregate measure does not ensure reliable performance on individual cases where predictions are highly conflicting.
Therefore, it is necessary to intentionally search for models of comparable statistical accuracy and assess prediction conflict as a notion of individual performance. To this end, one requires a technical measure of the degree of conflict between predictions. Such a conflict ratio will measure the amount  of conflicting predictions within a set of models of comparable statistical accuracy. 
While we condition on statistical accuracy here, the framework can readily be extended to condition on other group-level performance metrics or fairness notions. Therefore, our metrics are intended to complement existing approaches.

In the following Section \ref{sec:technical_section}, we discuss how a conflict ratio can be obtained from a technical perspective.

\begin{tcolorbox}[title={Summary of the legal analysis of the EU AI Act}, colback=black!05, colframe=black!60]
\begin{enumerate}[leftmargin=0.5cm]
    \item Articles 11, 13, and 15 of the AI Act in combination with Annex IV require providers of high-risk AI systems to inform about \textbf{performance for specific individuals.}
    \item {\bf Predictive multiplicity is a suitable framework} to measure individual performance and one possible way to satisfy the transparency and accuracy provisions with respect to specific persons. 
    \item Specifically, we recommend providers to enable deployers to assess individual performance \textbf{via conflict ratios for all affected individuals.}
\end{enumerate}
\end{tcolorbox}

\section{How to measure and evaluate predictive multiplicity} 
\label{sec:technical_section}

So far, we have argued that the EU AI Act requires performance evaluations at the individual level, and that measuring the conflicting predictions between models of comparable statistical accuracy can serve this purpose. 
In this section, we formally introduce the \textit{conflict ratio} as well as the \textit{$\delta$-ambiguity} as technical tools to quantify the disagreement between models for individual data points. 
Based on several experiments, we also provide an ad-hoc approach for how model developers can approximate the set of comparably good models in practice. 

\subsection{The Rashomon set, conflict ratio, and $\delta$-ambiguity}\label{sec:conflict_measures}
We consider the basic setup of binary classification on a dataset $\mathcal{D}$ that lies in $\mathcal{X} \times \mathcal{Y} \subseteq \R^{d} \times \{0, 1\}$. Models are trained on a training set $\mathcal{D}_{\mathrm{train}}$ and evaluated on a test set $\mathcal{D}_{\mathrm{test}}$ by computing the statistical accuracy $acc(g) = \frac{1}{\lvert \mathcal{D}_{\mathrm{test}}\rvert} \sum_{(x, y) \in \mathcal{D}_{\mathrm{test}}} \Indicator[g(x) = y]$. The class of all possible binary models is denoted by $\mathcal{H}_{\textrm{bin}} \coloneqq \{g: \mathcal{X} \to \{0, 1\}\}$ \citep{shalev2014understanding}. 
We follow \citet{Marx_2020_PredictiveMultiplicity} and define the Rashomon set relative to a fixed classifier $g_0 \in \mathcal{H}$ that achieves high statistical accuracy:

\begin{definition}[Rashomon set]
\label{def:rashomon_set}
    For a model class $\mathcal{H} \subseteq \mathcal{H}_{\textrm{bin}}$, a fixed classifier $g_0 \in {\arg\max}_{g \in \mathcal{H}}\hphantom{c}acc(g)$, and a threshold $\varepsilon \in [0, 1]$, the Rashomon set is defined as 
    \begin{align*}
        \mathcal{R}(\mathcal{H}, g_0, \varepsilon) \coloneqq \{ g\in \mathcal{H} \vert acc(g) \geq acc(g_0) - \varepsilon \}.
    \end{align*}
\end{definition}

Intuitively, the Rashomon set is the set of all models whose statistical accuracy is within an $\varepsilon$-range of the best possible classifier. 

\paragraph{\bf The conflict ratio measures conflict at the individual level}
To measure the degree of conflicting predictions for an individual case, we suggest using the conflict ratio. The conflict ratio can help deployers of high-risk AI systems identify individuals for whom reliable predictions cannot be made.

\begin{definition}[Conflict ratio]
For a data point $x \in \mathcal{X}$ and a Rashomon set $\mathcal{R}$, let $n_1(x, \mathcal{R}) \coloneqq \sum_{g \in \mathcal{R}} g(x)$ be the number of classifiers in $\mathcal{R}$ that predict label $1$, and $n_0(x, \mathcal{R}) \coloneqq \lvert \mathcal{R} \rvert - n_1$ be the number of classifiers in $\mathcal{R}$ that predict label $0$. 
The conflict ratio is defined as
    \begin{align*}
        c_{\mathcal{R}}(x) \coloneqq 
        \frac{\min\left\{ n_1(x, \mathcal{R}), n_0(x, \mathcal{R}) \right\} }
        {\lvert \mathcal{R} \rvert }
        \in [0, 0.5].
    \end{align*}
\end{definition}

Given a Rashomon set, one can assess the amount of conflict between its models' predictions on any data instance, be it from the training or test set, and even for new unseen and unlabelled data during deployment. The maximum value of $0.5$ is achieved if half of the models in $\mathcal{R}$ predict $1$ and the other half predict $0$. The minimal value of $0$ is achieved if all classifiers predict the same label. 

\paragraph{\bf $\delta$-Ambiguity measures conflict at the dataset-level.}
To measure the overall amount of conflicting predictions, we introduce the notion of $\delta$-ambiguity. Unlike the conflict ratio, which measures conflict at the individual level, the $\delta$-ambiguity quantifies the overall level of conflict in the dataset. The $\delta$-ambiguity can help deployers of high-risk AI systems to (a) assess the overall number of individuals with conflicting predictions and (b) assess the expressiveness of a Rashomon set. 
This metric will not be used to make case-based decisions.%

\begin{definition}[$\delta$-Ambiguity]
    For a threshold $\delta \in [0, 0.5]$, the $\delta$-ambiguity of a Rashomon set $\mathcal{R}$ on a dataset $\mathcal{D}$ is defined as
    \begin{align*}
        \mathcal{A}_{\delta}(\mathcal{R}, \mathcal{D}) 
        \coloneqq \frac{1}{\lvert \mathcal{D} \rvert} \sum_{x \in \mathcal{D}} \Indicator[c_{\mathcal{R}}(x) > \delta]
        \in [0, 1].
    \end{align*}
\end{definition}

Intuitively, the $\delta$-ambiguity is the fraction of individuals whose conflict ratio is at least $\delta$. 
It improves over the previous notion of ambiguity \citep{Marx_2020_PredictiveMultiplicity} --- from here on called \textit{standard ambiguity} --- in its ability to account for individual-level conflict: While standard ambiguity only checks whether or not a data point has a conflicting prediction, $\delta$-ambiguity checks if a data point reaches a certain level of conflict. 
Setting $\delta = 0$ in the above definition recovers standard ambiguity.

\subsection{How to approximate the set of comparably good models}\label{sec:ad-hoc}
To compute the conflict ratio and $\delta$-ambiguity, we need a Rashomon set of comparably good models. 
This Rashomon set should be as expressive as possible, that is, it should convey the relevant conflicting predictions. Unfortunately, an exhaustive search for all comparably good models is in general computationally infeasible and only solved for specific model classes \citep{Xin_2022_ExploringTheWholeRashomonSet, Marx_2020_PredictiveMultiplicity}. 
We therefore settle for an approximation that captures the most important factors from which predictive multiplicity originates during model development. 

\paragraph{\bf Dataset- and parametric multiplicity.} The model development process involves many decisions \citep{Suresh_2021_AFrameworkForUnderstanding, DAmour_2022_UnderspecificationPresents, Black_2024_TheLegalDuty}: Which dataset, which features, which model class, or which hyperparameters to use. 
All of these decisions affect the final model, and different decisions may yield conflicting predictions. 
We are going to separate decisions related to {\it dataset multiplicity} --- conflicting predictions due to variations in the data domain \citep{GeorgeNeedellUstun_2025_ObservationalMultiplicity, Cooper_2024_ArbitrarinessAndSocialPrediction, Meyer_2023_TheDatasetMultiplicityProblem} --- and {\it parametric multiplicity} --- conflicting predictions due to variations in the model domain \citep{Kulynych_2023_ArbitraryDecisions, HsuCalmon_2022_RashomonCapacity, Marx_2020_PredictiveMultiplicity}. As we show in Section \ref{sec:experiment_dataset_multiplicity}, accounting for both dataset and parametric multiplicity is crucial for the expressivity of the Rashomon set.  

\paragraph{\bf The ad-hoc approach for constructing Rashomon sets.} A practical approach to approximate the Rashomon set for any model class is the ad-hoc approach \citep{Marx_2020_PredictiveMultiplicity,Kulynych_2023_ArbitraryDecisions, Cooper_2024_ArbitrarinessAndSocialPrediction}. 
Given the model class $\mathcal{H}$ and a dataset $\mathcal{D}$, we find the best performing model $g_0$ via cross-validation. 
We then train many different models, taking into account both dataset multiplicity (for example, by bootstrapping the data) and parametric multiplicity (for example, by varying hyperparameters or using different random seeds). 
The Rashomon set then includes all models with high statistical accuracy (see Definition \ref{def:rashomon_set}).

\subsection{The importance of dataset multiplicity in evaluations}
\label{sec:experiment_dataset_multiplicity}

In this section, we present two experiments --- one on synthetic data and one on real-world data from the American Community Survey (ACS) --- that highlight the importance of dataset multiplicity. 

\paragraph{\bf Technical Methodology} We train decision trees and explore parametric multiplicity by systematically varying tree depth, splitting criterion, and random seed. For this purpose, we build a grid of possible parameter combinations and randomly select a combination without replacement per model. This grid, and with it the integration of parametric multiplicity, is fixed for all experiments. In contrast, we vary the degree to which we integrate dataset multiplicity, from \textit{no dataset multiplicity} to \textit{full dataset multiplicity}. We test different approaches such as bootstrapping, adding feature noise and feature subsampling and assess their effectiveness. For further details see Supplement~\ref{App:details_exp_data_mult}.
 
\paragraph{\bf Illustrative example on synthetic data}
To illustrate the importance of dataset multiplicity, we first present a controlled experiment on synthetic data. 
The advantage of this setup is that the true individual conflict ratios are known by construction. 
To form the dataset, we draw data points from a set of 2D Gaussian distributions:  $\mathcal{N}((0,0), (1,1))$ and $\mathcal{N}((5,5), (1,1))$ with label $0$; and $\mathcal{N}((5,5), (1,1))$ and $\mathcal{N}((10,10), (1,1))$ with label $1$. 
The two Gaussians with mean vector $(5,5)$ completely overlap with mixed labels, meaning that the data points in this region will have a true conflict ratio of $0.5$.
All remaining data points have a ground truth conflict ratio of $0$. 
We explore four different approaches to approximate the Rashomon set by systematically varying the amount of dataset multiplicity:
\begin{itemize}
    \item[(a)] {\bf No dataset multiplicity:} We use the same training data sample of $8,000$ points for all $10,000$ models.
    \item[(b)] {\bf Bootstrapped dataset multiplicity on few models:} We train $200$ models on bootstrapped training data, which is created by drawing $800$ points with replacement from the original $8,000$ training points.
    \item[(c)] {\bf Bootstrapped dataset multiplicity on many models:} We train $10,000$ models on bootstrapped training data similar to (b).
    \item[(d)] {\bf Full dataset multiplicity:} We draw an entirely new dataset of $8,000$ training samples for each of the $10,000$ models.
\end{itemize}
The best classifier with test accuracy $acc(g_0)\approx 0.782$ was found by (d).
Together with $\varepsilon = 0.1$ this determines the Rashomon set threshold (see Definition~\ref{def:rashomon_set}), which we keep fixed for all approaches.

\begin{figure}[t]
    \centering
    \includegraphics[width=\textwidth]{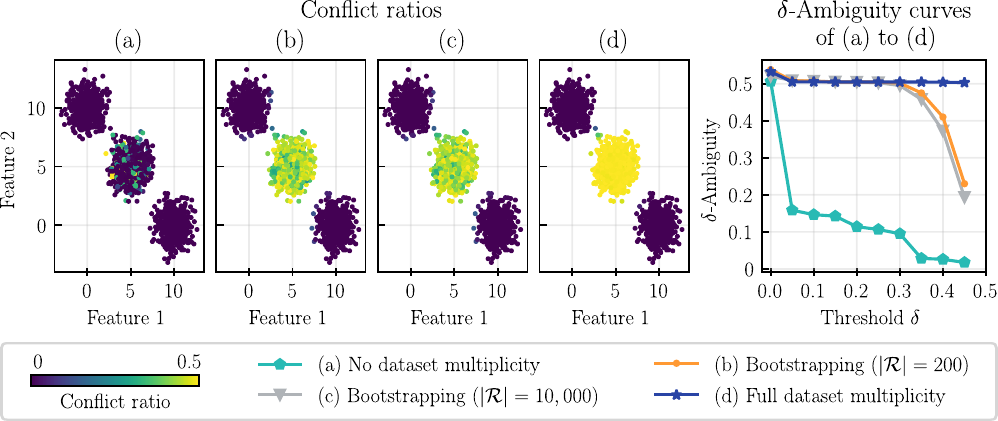}  
    \caption{\textbf{The importance of dataset multiplicity motivated on two-dimensional synthetic data.} For fixed parametric multiplicity, we compare different ways to include dataset multiplicity into the ad-hoc approach from (a) no dataset multiplicity to (d) full dataset multiplicity. 
    The ability to discover conflicting data points in the overlap region $\mathcal{N}((5, 5), (1,1))$ increases with stronger integration of dataset multiplicity (scatter plots; left to right).
    The $\delta$-ambiguity curves (right plot) depict the extent of the improvement for different conflict ratio thresholds $\delta$.
    The ground truth would be a constant $0.5$ curve.
    }
    \label{fig:simulateGaussian} 
\end{figure}

Figure~\ref{fig:simulateGaussian} depicts the impact of the choice of dataset multiplicity on the estimation of conflict ratios and $\delta$-ambiguities. 
Both measures are evaluated on a test set of $2,000$ data points.
By construction, we know that half of the data points, located in the middle Gaussian, have ground truth conflict ratio $0.5$, while the other half have conflict ratio $0$.
A $\delta$-ambiguity curve of constant $0.5$ would represent an optimal Rashomon set that finds ground truth conflict ratios.
We see that accounting for dataset multiplicity is crucial: The \textit{no dataset multiplicity} condition (a) leads to a Rashomon set of models that behave similarly even in the overlap region. 
As a consequence, this condition significantly underestimates predictive multiplicity both at the individual and at the dataset level: We see that the $\delta$-ambiguity curve is significantly below the other curves (turquoise line, right plot). 
On the other hand, \textit{full dataset multiplicity} (d) provides a diverse Rashomon set and is able to recover the ground truth conflict ratios. 
Interestingly, we observe that the \textit{bootstrapped dataset multiplicity} methods (b and c) --- which are practical given that drawing bootstrap samples is something that we can do in practice even for smaller datasets --- already provide a good approximation of the ground truth conflict ratio, even if we train only 200 models (b). 
Here, the difference to the \textit{full dataset multiplicity} only becomes apparent for high values of $\delta$ (right plot).

\paragraph{\bf Insights from illustrative example translate to real-world data}
After motivating the importance of dataset multiplicity in a controlled setting, we now demonstrate it on data from the {\bf American Community Survey (ACS)} \citep{Ding_2021_RetiringAdult}. 
In this setup, we lack knowledge of the ground truth conflict ratios. 
Hence, we can only compare the approaches and their resulting Rashomon sets with each other.
For two Rashomon sets $\mathcal{R}_1$ and $\mathcal{R}_2$, and a fixed test dataset $\mathcal{D}$ we calculate the distance in conflict ratios, that is, 
\begin{align*}
    Dist_{\mathcal{D}}(\mathcal{R}_1, \mathcal{R}_2) \coloneqq
    \frac{1}{\lvert \mathcal{D} \rvert} 
    \sum_{x \in \mathcal{D}} \lvert c_{\mathcal{R}_1}(x) - c_{\mathcal{R}_2}(x) \rvert
    \in [0, 0.5].
\end{align*}
If we assumed that the conflict ratios were uniformly distributed across the interval $[0, 0.5]$, the expected distance would be $1/6$. Two approaches would then be similar if their distance is less than $1/6$.
Additionally to the distance, we look at the standard ambiguity $\mathcal{A}_{\delta=0}(\mathcal{R}, \mathcal{D})$ and the $0.4$-ambiguity $\mathcal{A}_{\delta=0.4}(\mathcal{R}, \mathcal{D})$ to judge the diversity of the models present in a Rashomon set $\mathcal{R}$. 

We use the ACSEmployment prediction task, which determines whether an individual is employed.
After preprocessing (see Supplement~\ref{App:data_preprocessing}), the data has $13$ features and $\approx 1.7$ Million data points from which we extract a test sample of $7,000$ points.
We found that this sample size suffices to achieve statistical accuracy beyond the saturation point (see Supplement~\ref{App:accuracy_convergence}).

Again, we keep parametric multiplicity fixed: We train decision trees and systematically vary the depth, splitting criterion, and random seed.
In Supplement~\ref{App:additional_experiments_ACS} and \ref{App:additional_experiments_further_datasets}, we include further experiments on different model classes and European datasets relevant to the high-risk provisions of the AI Act.
We extend the list of data variations from above and compare the following approaches to incorporate dataset multiplicity:
\begin{enumerate}
    \item[(a)] {\bf No dataset multiplicity:} We train $253$ models, all on the same fixed training data sample of $7,000$ points.
    \item[(b, c)] 
    {\bf Bootstrapping:} We train $253$ models on bootstrapped training data, which is created by drawing $2,800$ (b) or $4,200$ (c) points with replacement from a fixed training data sample of $7,000$ points.
    \item[(d)] {\bf Feature subsampling:} We use a fixed training data sample of $7,000$ points for all $253$ models. Each model is trained on a random feature subset of $12$ out of the $13$ features.
    \item[(e)] {\bf Feature noise:} 
    We use a fixed training data sample of $7,000$ points and train $253$ models, each on a noisy version.
    For numerical columns we add normally distributed noise of standard deviation $0.05$. 
    For categorical columns we randomly switch the category with probability $0.001$.
    The values are chosen in a way to maximise statistical accuracy.
    Too much noise leads to less accurate models that do not meet the Rashomon set threshold.
    \item[(f)] {\bf Full dataset multiplicity:} We train $253$ models each on an entirely new dataset of size $7,000$. This is possible since the ACS dataset is large enough. 
\end{enumerate}

The best classifier with test accuracy $acc(g_0)\approx 0.776$ was again found by the approach of \textit{full dataset multiplicity} (f).
Similar to the illustrative example above, we keep the Rashomon set threshold with $\varepsilon = 0.1$ fixed for all approaches.

\begin{figure}[t]
    \centering
    \includegraphics[width=\textwidth]{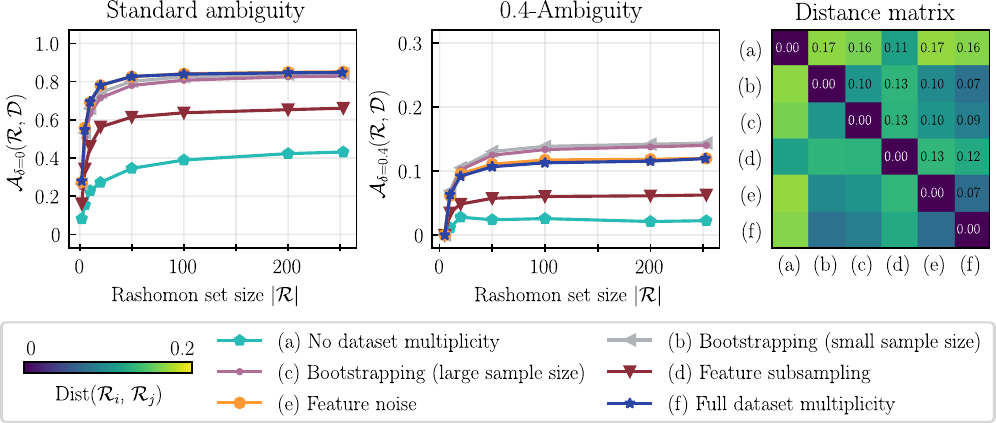}  
    \caption{
    \textbf{The importance of dataset multiplicity on ACSEmployment data.}
    For fixed parametric multiplicity, we compare different ways to include dataset multiplicity into the ad-hoc approach from (a) no dataset multiplicity to (f) full dataset multiplicity.
    The integration of dataset multiplicity increases the ability to discover conflicting data points (left plot) and highly conflicting data points (middle plot). 
    The effectiveness differs between approaches.
    Approaches that recover standard and 0.4-ambiguity well (left and middle plot) also produce similar individual conflict ratios with small pairwise distances (right plot).
    }
    \label{fig:ACSEmployment_DT} 
\end{figure}

Figure~\ref{fig:ACSEmployment_DT} displays the results of the experiment. 
The standard ambiguity $\mathcal{A}_{\delta=0.0}(\mathcal{R}, \mathcal{D})$ (left plot) and the $0.4$-ambiguity $\mathcal{A}_{\delta=0.4}(\mathcal{R}, \mathcal{D})$ (middle plot) are plotted against the size of the Rashomon set.
On the right, the distances $Dist_{\mathcal{D}}(\mathcal{R}_i, \mathcal{R}_j)$ between all approaches (a to f) are depicted as a heat map. 
From Figure \ref{fig:ACSEmployment_DT}, we again see that ignoring dataset multiplicity completely (turquoise line) yields the poorest performance in terms of standard and $0.4$-ambiguity. 
The resulting Rashomon set falls behind in terms of its ability to find conflicting (left plot) and highly conflicting individuals (middle plot). 
It also shows higher distance in conflict ratios to the better performing approaches (right plot), with the exception of feature subsampling.
However, feature subsampling also falls behind in terms of standard and $0.4$-ambiguity.
The remaining approaches perform similarly well and strongly agree in their individual conflict ratios, that is, they have smaller pairwise distances (right plot). %
Interestingly, both bootstrapping methods exceed even the full dataset multiplicity condition in terms of $0.4$-ambiguity.

In summary, we find that incorporating dataset multiplicity in any way significantly improves the diversity of Rashomon sets obtained by the ad-hoc approach.
Between the different ways of enforcing data variations, bootstrapping seems to dominate.
For a generalisation study with more datasets and model classes see Supplement~\ref{App:additional_experiments_ACS}.

\subsection{The ad-hoc approach closely approximates the true Rashomon set}\label{sec:experiment_treefarms}

\begin{figure}[b]
    \centering
    \includegraphics[width=0.9\textwidth]{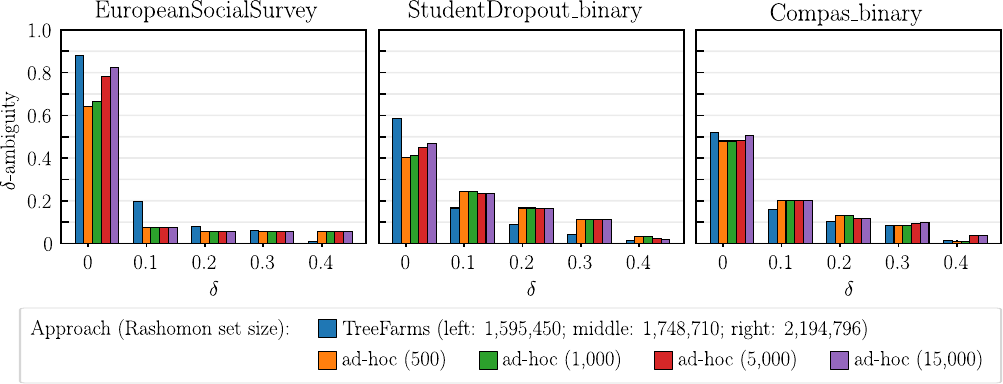}  
    \caption{
    \textbf{Comparison of the ad-hoc approach and TreeFarms \citep{Xin_2022_ExploringTheWholeRashomonSet}.}
    For three different datasets (details in Table~\ref{table:datasets} in Supplement~\ref{App:data_preprocessing}) we use the ad-hoc approach to train Rashomon sets of different sizes and track the $\delta$-ambiguities.
    Results are compared to the ground truth Rashomon set found by TreeFarms \citep{Xin_2022_ExploringTheWholeRashomonSet}. For all datasets we use a 80-20 train-test split.
    }
    \label{fig:Xin_comparison} 
\end{figure}

In this section, we investigate the ability of the ad-hoc approach to approximate the true Rashomon set.
For that purpose, we experimentally compare the ad-hoc approach to the TreeFarms algorithm proposed by \citet{Xin_2022_ExploringTheWholeRashomonSet}. 

\paragraph{\bf Technical Methodology}
The TreeFarms algorithm is a method to fully compute the Rashomon set for sparse decision trees on binary datasets. 
In the experiment, we calculate Rashomon sets using the ad-hoc approach and TreeFarms, and track the $\delta$-ambiguities. 
In TreeFarms, sparsity is enforced by a penalty on the number of leaf nodes. 
For comparability, we slightly modify the ad-hoc approach in the same way (details in Supplement~\ref{App:details_exp_treefarms}).
We perform the experiment on binarised versions of compas \citep{COMPAS, Hu_2019_OptimalSparseDecisionTrees}, studentDropout \citep{dataset_studentDropout}, and a dataset retrieved from the European Social Survey \citep{ESS_data}. The latter two are from Europe and could potentially be used to train high-risk AI systems.

\paragraph{\bf Results}
Figure~\ref{fig:Xin_comparison} displays the $\delta$-ambiguities for the different Rashomon sets that we find via TreeFarms and the ad-hoc approach.
For all three datasets, the standard ambiguity of the ad-hoc approach underestimates that of TreeFarms.
This is as expected, since standard ambiguity is monotone: If new models are added to the Rashomon set, its value can only increase.
Interestingly, we see that even for comparably small Rashomon sets the standard ambiguity of the ad-hoc approach gets close to that of TreeFarms.
For example in the case of compas, with only $\lvert \mathcal{R} \rvert = 500$ models the ad-hoc approach reaches a standard ambiguity of $\mathcal{A}_{\delta=0}(\mathcal{R}, \mathcal{D}) \approx 0.4799$. 
TreeFarms with a Rashomon set $\mathcal{R}$ of over $2$~Million models yields a standard ambiguity of $\mathcal{A}_{\delta=0}(\mathcal{R}, \mathcal{D}) \approx 0.5204$.

For $\delta > 0$, the $\delta$-ambiguity is no longer monotone. 
Hence, the ad-hoc approach could both over- or underestimate the ground truth $\delta$-ambiguities of TreeFarms. While we observe an overestimation for studentDropout\_binary and compas\_binary, we tend to underestimate for europeanSocialSurvey.
It is not surprising to see some deviations here. 
Unlike the ad-hoc approach, TreeFarms assumes that each model in the Rashomon set is equally likely.
The ad-hoc approach instead mimics usual training procedures.
It therefore tends to produce models that would actually occur in practice, leaving the unlikely models aside.

The key takeaway is that the ad-hoc approach seems to perform well in terms of its ability to detect conflicting and highly conflicting individuals for decision trees. 
Deviations from to the ground truth of TreeFarms are rather small.

\subsection{Small, diverse Rashomon sets can go a long way}\label{sec:small_RS}
In the previous Sections \ref{sec:experiment_dataset_multiplicity} and \ref{sec:experiment_treefarms}, we have seen that even small Rashomon sets estimate conflict ratios and $\delta$-ambiguities surprisingly well. 
Specifically, in our experiments on ACSEmployment (Figure~\ref{fig:ACSEmployment_DT}), standard and $\delta$-ambiguities converge very early in terms of the Rashomon set size, suggesting that a small Rashomon set suffices to be representative.
It is not by a high number of models that diversity is obtained, but through the combined effects of dataset and parametric multiplicity. 
Similarly, in Figure~\ref{fig:Xin_comparison}, we have seen that already small Rashomon sets retrieved by the ad-hoc approach can suffice to closely approximate the $\delta$-ambiguities of TreeFarms with roughly $2$ Million models. 

Taken together, our findings suggest that practitioners can obtain reasonably good approximations of predictive multiplicity using a relatively small number of models, provided they account for all important factors of multiplicity. 
\begin{tcolorbox}[title={Summary of the technical approaches to measure individual performance}, colback=black!05, colframe=black!60]
\begin{enumerate}[leftmargin=0.5cm]
    \item In order to comply with the legal requirements of the AI Act, we suggest \textbf{conflict ratios} to measure  individual performance and  \textbf{$\delta$-ambiguity} as an aggregate assessment on the dataset level. 
    \item To compute these measures, it often suffices to use small Rashomon sets, as long as they are based on both parametric and dataset multiplicity. 
\end{enumerate}
\end{tcolorbox}

\subsection{Limitations}\label{Sec:limitations}
Central to this work is the ad-hoc approximation of the Rashomon set. As this approach is non-exhaustive, the resulting set remains a subset of the true Rashomon set. Hence, while a conflict ratio $> 0$ is sufficient to identify a conflicting data point, a zero conflict ratio should be interpreted more cautiously: It may indicate a non-conflicting data point, but it may also point to a lack of expressivity of the Rashomon set. However, in our empirical evaluation on sparse decision trees (Section~\ref{sec:experiment_treefarms}) we are able to identify most of the conflicting data points. Future research on exhaustive methods might reveal better alternatives, but for now the ad-hoc approach is the only computationally feasible and model agnostic option.
Our analysis further assumes a fixed model class. Since Rashomon sets depend on the choice of model class, conflict ratios should be interpreted relative to the chosen class. Selecting the model class is a key decision for practitioners, who often balance predictive performance with other considerations such as interpretability, for example, by choosing to work with decision trees.

\section{Discussion}
\label{sec:Discussion}

In this paper, we argue that predictive multiplicity has significant implications for the EU AI Act's accuracy provisions for high-risk AI systems (Art.~15 AIA). 
We present a legal analysis of the AI Act and its accuracy provisions followed by a computational analysis of measuring individual performance under predictive multiplicity. 

\paragraph{\bf Recommendation: Integrating predictive multiplicity considerations for providers and deployers}
We observe that the AI Act mandates providers of high-risk AI systems to examine individual performance under its provisions. 
Ultimately, it is the harmonised standards that will govern how performance needs to be assessed in practice (Art.~40(1) AIA) \citep[Art.~40, mn.~1 and 2]{Martini_2024_Art40} \citep{gorywoda2009theneweuropean, kilian2025european}. However, the standards are still forthcoming \citep{JTC21_2025_WorkProgramme, EU_2023_StandardisationRequest, EU_2025_StandardisationRequest} and their focus on individual performance remains unclear.
Predictive multiplicity underscores that any AI model is always only one of several viable options. 
We argue that if models of comparably high statistical accuracy diverge significantly on individual data points, then they are not capable of making a meaningful decision on that data point. 
Hence, we make the following recommendations:

\begin{tcolorbox}[title={Recommendations for addressing predictive multiplicity within the AI Act}, colback=black!05, colframe=black!60]
\begin{enumerate}[leftmargin=0.5cm]
    \item\label{recommendations:one} In order to comply with the AI Act, {\bf providers} of high-risk AI systems could offer tools that enable \textbf{deployers} to evaluate the amount of conflict between models in the Rashomon set on \textbf{all} individuals.
    \item As a best practice, {\bf deployers} of high-risk systems should use conflict ratios to identify cases of high ambiguity --- going beyond what the AI Act currently requires. For these cases, deployers should not solely rely on automated decision-making but involve human oversight.
\end{enumerate}
\end{tcolorbox}

\paragraph{\bf Recommendations extend previous works}
Previous research pointed out implications of predictive multiplicity for US law \citep{Laufer_2025_WhatConstitutes, Black_2024_TheLegalDuty}.
To our knowledge, we provide the first connection to the AI Act, and we formulate concrete, actionable recommendations.
Our focus on individual data points complements existing research on measuring predictive multiplicity per data point \citep{Cooper_2024_ArbitrarinessAndSocialPrediction, HsuCalmon_2022_RashomonCapacity} with two straightforward and easy-to-communicate measures: conflict ratio and $\delta$-ambiguity. 
The bottleneck of their calculation --- as for other existing measures --- is the accessibility of the Rashomon set of comparably accurate models.
Methods that exhaustively enumerate the Rashomon set do exist, but are restricted to specific model classes \citep{Xin_2022_ExploringTheWholeRashomonSet, Marx_2020_PredictiveMultiplicity}.
We overcome this limitation by suggesting an ad-hoc approach that systematically exploits parameter variations and dataset variations to get an approximation of the Rashomon set.
In comparison to exhaustive methods, the ad-hoc approach is easy to implement, computationally light, and most importantly, applicable to any model class. 
Our experiments reveal that it is in particular variations in the data domain that enable the ad-hoc approach to find expressive Rashomon sets (Section~\ref{sec:experiment_dataset_multiplicity}), which is in line with previous research \citep{Meyer_2023_TheDatasetMultiplicityProblem, GeorgeNeedellUstun_2025_ObservationalMultiplicity}.
In summary, our computational results motivate a straightforward approach where model deployers could store a small number of alternative models of high statistical accuracy to be able to evaluate predictive multiplicity.

\paragraph{\bf Outlook}
Overall, we believe that predictive multiplicity warrants deeper consideration within the discourse surrounding the AI Act. 
In high-risk settings, decisions mediated by AI systems can have profound consequences on individuals' lives.
In this work, we demonstrate how predictive multiplicity --- a challenge inherent to the training of machine learning models --- can be used as a valuable tool for assessing individual-level performance. 
Continued investigation into predictive multiplicity and its implications for regulation will be crucial to ensure that the AI Act effectively addresses the complexities of modern AI-driven decision-making.

\section*{Generative AI usage statement}
The authors used generative AI tools for grammar checking, wording refinement, and assistance with code writing. No scientific content, arguments, or conclusions were generated by large language models.

\begin{acks}
We thank Mich\`{e}le Finck, Rabanus Derr, Mila Gorecki, Benedikt H\"{o}ltgen, Michelle Acevedo Callejas, and
Lucas G. Uberti-Bona Marin as well as our anonymous reviewers and area chair for their valuable feedback. This work has been supported by the German Research Foundation through the Cluster of Excellence ``Machine Learning -- New Perspectives for Science'' (EXC 2064/1 number 390727645) and DFG project 560788681, the Carl Zeiss Foundation through the CZS Institute for Artificial Intelligence and Law, and the International Max Planck Research School for Intelligent Systems (IMPRS-IS).
\end{acks}

\bibliography{references}

\newpage

\appendix

\vspace{2cm}
\centerline{\Huge Supplemental Material}

\section{Additional notes on AI categories regulated under the EU AI Act}\label{App:AI_categories}
As already pointed out in the main paper, the AI Act aims to strike a balance between human-centric, trustworthy AI and innovation and to improve the functioning of the internal Union market (Art.~1(1) AIA). It acknowledges that AI can bring diverse benefits at the individual, societal, and environmental level (Rec.~4 AIA), while also recognising that --- depending on the area of application --- it can cause harm (Rec.~5 AIA). To balance these benefits and potential harms, it takes a risk-based approach where it differentiates between four risk categories: prohibited, high-risk, transparency risks, and minimal to no risk. Depending on their risk level, different provisions apply to different kinds of AI systems (Rec.~26 AIA) \citep{EPRS_2025_AIAImplementationTimeline}. ``Risk'', in the understanding of the AI Act, is ``the combination of the probability of an occurrence of harm and the severity of that harm'' (Art.~3(2) AIA). Where AI systems come with a high risk, the goal is to bring this risk down to an acceptable level (Art.~9(5) AIA). Originally falling outside the scope of the AI Act's risk-based approach --- they were no part of the European Commission's initial draft but were added at the proposal of the Council and the Parliament --- the AI Act furthermore regulates general-purpose AI models (GPAIMs) \citep{EPRS_2025_AIAImplementationTimeline}. 
In the main paper we focus primarily on high-risk AI systems, which are subject to the majority of the AI Act's provisions. Here, we complement the picture by providing further information on the other AI categories regulated under the AI Act.

\paragraph{\bf Prohibited AI practices}
For some AI practices, the AI Act considers the risk to be unacceptable and therefore prohibits them. Due to their unacceptable risk, the provisions for prohibited AI practices have already applied since 2 February 2025 (Art.~113 AIA). These practices are listed in Article 5 of the AI Act and include harmful manipulation or deception, social scoring, criminal risk-assessments of natural persons, and real-time remote biometric identification (RBI), among others (Art.~5(1) AIA). Regarding the latter, however, concerns have been raised about the effectiveness of fundamental rights protection, especially due to the provision's narrow scope on real-time RBI (as opposed to post RBI) \citep{Giannini_2024_AIA}. 

\paragraph{\bf Transparency obligations for providers and deployers of certain AI systems}
The transparency obligations set out in Article 50 of the AI Act are motivated by the increasing availability of AI-generated content, which can be hard to distinguish from human-generated content. Targeting the providers and deployers of certain AI systems, including generative and interactive AI systems and deepfakes, they aim to ensure that natural persons are informed once they are interacting with AI or exposed to AI generated content, thereby mitigating the risk of misinformation, fraud, or consumer deception \citep{Commission_2025_GuidelinesTransparentAISystems}. The transparency obligations will apply from 2 August 2026 (Art.~113 AIA).

\paragraph{\bf General-purpose AI models}
A general-purpose AI model (GPAIM) is defined as an AI model ``that displays significant generality and is capable of competently performing a wide range of distinct tasks [\dots]'' (Art. 3(63) AIA). This includes models ``trained with a large amount of data using self-supervision at scale'' (Art. 3(63) AIA). Most prominent examples of such GPAIMs are large language models. The AI Act refines between GPAIMs and GPAIMs with systemic risk, where the systemic risk emerges from a GPAIM having ``high-impact capabilities'', that is, ``capabilities that match or exceed the capabilities recorded in the most advanced general-purpose AI models'' (Art.~3(64) AIA). Such capabilities might become a systemic risk where they have a ``significant impact on the Union market due to their reach, or due to actual or reasonably foreseeable negative effects on public health, safety, public security, fundamental rights, or the society as a whole, that can be propagated at scale across the value chain'' (Art.~3(65) AIA). Where a GPAIM --- including one with systemic risk --- additionally falls under the high-risk categories, provisions for GPAIMs (with systemic risk) and high-risk AI systems both apply (Art.~53(1)(b)(i) AIA; Rec.~85 and Rec.~97). Notably, while the rest of the AI Act targets AI systems, the GPAIM provisions address AI models. This reflects that GPAIMs are products on their own, which may subsequently be integrated into various AI systems. The provisions for GPAIMs became effective on 2 August 2025, except for Article 101 of the AI Act regulating the fines that can be imposed on providers of general-purpose AI models in case of non-compliance (Art.~113 AIA). To support providers of GPAIMs in demonstrating compliance, the European Commission published the general-purpose AI code of practice (Art.~56 AIA) \citep{Commission_2025_GPAIMCode}. The code is a voluntary tool that providers can sign to publicly show they comply with the AI Act by adhering to it.

\paragraph{\bf Minimal to no risk} AI systems that fall under none of the AI categories above and are not considered high-risk remain mostly unregulated under the AI Act. The only exception is that providers and deployers of AI systems need to make sure that persons operating with AI systems on their behalf are sufficiently ``AI literate'', that is, they have been equipped with the necessary knowledge to make informed decision regarding AI systems, depending on the relevant context of use (Art.~4 AIA, Rec.~20). As part of the general provisions of the AI Act, Article 4 applies since 2 February 2025.

\section{Technical aspects of predictive multiplicity}
\label{app:why_multiplicity}
In the main paper, we have argued that the AI Act requires providers of high-risk AI systems to inform about performance for specific individuals, and that measuring predictive multiplicity presents
a suitable approach to comply with these provisions. Here, we provide an additional discussion of why we believe that predictive multiplicity is an appropriate technical tool to measure individual performance.

\paragraph{\bf Predictive multiplicity can be studied in all modelling pipelines} A desirable property of predictive multiplicity is that it can be studied in all machine learning pipelines, without making assumptions \citep{DAmour_2022_UnderspecificationPresents}. This is particularly important in the context of broad regulations like the AI Act, which may apply to many different machine learning setups. What is more, the results of an analysis via predictive multiplicity are relatively straightforward to interpret: Re-training with a different parameter setup on a different subset of the data would produce a different model and different predictions on certain individuals. We argue that this is important in social contexts where model developers, regulators, and the affected individuals may have different incentives \citep{bordt2022post}.

\paragraph{\bf The relationship between predictive multiplicity and uncertainty estimation} Uncertainty estimation methods \citep{hullermeier2021aleatoric} are another technical approach to assess whether predictions for a specific individual are reliable. In some settings, for example when uncertainty arises from noise in the data and the model class is well-specified, the results of uncertainty quantification and analysis via predictive multiplicity may agree \citep{Watson-Daniels_2023,HsuCalmon_2022_RashomonCapacity,Ganesh_2025_TheCuriousCase}. In general, however, different calibrated models may produce conflicting predictions, and the precise nature of the relationship between uncertainty quantification and predictive multiplicity is a topic of ongoing research \citep{roth2023reconciling,roth2025resolving,hullman2025multiplicity}. An analysis via predictive multiplicity, also for calibrated models, may be especially important when there is a distribution shift between training and deployment \citep{DAmour_2022_UnderspecificationPresents}.

\section{Datasets}\label{App:data_preprocessing}
This section describes the datasets used in our experiments and their preprocessing. All datasets are representative of high-risk AI use cases --- that is, if used to train AI systems, they would likely fall under the AI Act's high-risk category.

\begin{table}[H]
\begin{tabular}{|l|l|l|l|l|l|l|}
\hline
\bfseries dataset & \bfseries area & \bfseries target & \bfseries \#data points & \bfseries \#features & \bfseries chance acc.& \bfseries source \\ \hline\hline
ACSEmployment & Employment & employed or not & $1,782,307$ & $13$ & $0.5313$ & \citep{Ding_2021_RetiringAdult}   \\ \hline
ACSIncome & Finance & earns more than 50k or not & $195,665$ & $10$ & $0.5894$ & \citep{Ding_2021_RetiringAdult}   \\ \hline
adult& Finance & earns more than 50k or not & $23,374$ & $10$ & $0.5$ & \citep{adult} \\ \hline
creditCardsClients & Finance & will default or not & $13,042$ & $23$ & $0.5$ & \citep{dataset_CreditCardsClients}   \\ \hline
europeanSocialSurvey& Finance & earns more than median & $1,107$ & $20$ & $0.5285$ & \citep{ESS_data} \\
& &income in dataset or not & &  &  & \\ \hline
southGermanCredit & Finance & high credit risk or not & $600$ & $14$ & $0.5$ & \citep{southGermanCredit_data}    \\ \hline
compas& Recidivism & will reoffend or not & $6,907$ & $14$ & $0.5066$ & \citep{COMPAS}\\ \hline
compas\_binary& Recidivism & will reoffend within 2 years& $6,907$ & $12$ & $0.5372$ & \citep{COMPAS, Hu_2019_OptimalSparseDecisionTrees} \\ 
& & or not & &  &  & \\ \hline
studentPerformance& Education & will get high grade or not & $649$ & $27$ & $0.5362$ &\citep{dataset_portugueseStudent} \\ \hline
oulad& Education & student will pass or not & $28,657$ & $7$ & $0.582$ & \citep{dataset_oulad} \\ \hline
studentDropout& Education & student will drop out or not & $2,828$ & $29$ & $0.5$ & \citep{dataset_studentDropout} \\ \hline
studentDropout\_binary & Education & student will drop out or not & $2,828$ & $19$ & $0.5$ & \citep{dataset_studentDropout} \\ \hline
breastCancer& Medicine & has breast cancer or not & $424$ & $30$ & $0.5$ & \citep{breastCancer} \\ \hline
\end{tabular}
\caption{Datasets used in various experiments throughout the paper. Imbalanced datasets (chance accuracy > 0.6) were balanced to comprise equally many data points with labels $0$ and~$1$, that is, chance accuracy of $0.5$. Categorical features are either binarised or deleted.}
\label{table:datasets}
\end{table}

\paragraph{\bf ACSEmployment}
The ACS data can be accessed via the folktables package of \citet{Ding_2021_RetiringAdult} for Python. To download the data, we used:
\begin{verbatim}
    survey_year="2018", horizon="5-Year", survey="person", states=["CA"]
\end{verbatim} 
Three features were dropped and some of the categorical features were binarised (see Table~\ref{table:ACS_preprocessing}).
For the remaining features we deleted all individuals with missing values, which removed 97,616 out of originally 1,879,923 data points.
Numerical features (AGEP), binary features (SEX, DIS, NATIVITY, DEAR, DEYE, DREM) and ordinal features (SCHL) were not modified.
The final dataset has 13 features and 1,782,307 data points.
\begin{table}[h]
\begin{tabular}{|l|p{4cm}|p{9cm}|}
\hline
\bfseries feature & \bfseries description  & \bfseries modification \\ \hline\hline
ESP & Employment status parents  & Dropped, because it is only recorded for persons under the age of 18, a minority in the ACSEmployment prediction task.\\ \hline
ANC & Ancestry recode  & Dropped, because this feature just tells if a person gave information on their ancestry/ethnic background (not which!).\\ \hline
MIL & Military service  & Dropped, because the vast majority of persons recorded in the ACS data never served in military. \\ \hline
MAR    & Marital status & Binarised to: not married vs. married         \\ \hline
CIT    & Citizenship status & Binarised to: not US citizen vs. US citizen   \\ \hline
MIG    & Mobility status & Binarised to: non-movers vs. movers\\ \hline
RAC1P  & Race & Binarised to: non-White vs. White \\ \hline
RELP   & Relationship to householder & Binarised to: not householder vs. householder \\ \hline
\end{tabular}
\caption{Dropped and binarised features for ACSEmployment data.}
\label{table:ACS_preprocessing}
\end{table}

\paragraph{\bf ACSIncome}
Similar to ACSEmployment, we used the folktables package of \citet{Ding_2021_RetiringAdult} for Python to download the data. We used the following configuration:
\begin{verbatim}
    survey_year="2018", horizon="1-Year", survey="person", states=["CA"]
\end{verbatim} 
The features were processed similar to ACSEmployment.

\paragraph{\bf adult}
To download the data, we used \verb|fetch_openml(name="adult", version=2, as_frame=True)| from the \verb|sklearn.datasets| package for Python.
We binarised the features `native-country', `marital-status', and `race' and balanced the dataset by subsampling to achieve a chance accuracy of $0.5$.

\paragraph{\bf creditCardClients} 
We binarised the feature `MARRIAGE' and kept only data points with values in $\{1,2,3,4\}$ of the feature `EDUCATION'. As target feature we used `default payment next month'. The dataset was balanced by subsampling to achieve a chance accuracy of $0.5$.

\paragraph{\bf europeanSocialSurvey}
The data of the European Social Survey can be accessed through the ESS Data Portal. We used the data of round 9 from 2018. We used the features [`nwspol', `stflife', `gndr', `agegroup', `edubde1', `grspnum', `emplrel', `infqbst']. All non-binary features were binned. The chance accuracy after dropping missing values, was $0.5285$.

\paragraph{\bf southGermanCredit}
We dropped some features that were not binary, ordinal, or numerical: [`verw', `famges', `buerge', `weitkred', `beruf', `pers'] and further binarised the features `wohn' and `miete'. As target we used the feature `kredit'. The dataset was balanced by subsampling to achieve a chance accuracy of $0.5$.

\paragraph{\bf compas and compas\_binary}
We deleted all features except [`sex', `age', `juv\_fel\_count', `decile\_score', `juv\_misd\_count', `juv\_other\_count', `priors\_count', `days\_b\_screening\_arrest', `c\_days\_from\_compas', `c\_charge\_degree', `start', `end', `event']. We further calculated the length of stay in jail in days. As target variable we used is\_recid. Finally, we dropped missing values. The final dataset was balanced with chance accuracy $0.5066$. For the binary version of compas, we used the dataset of \citep{Hu_2019_OptimalSparseDecisionTrees}, who uploaded their version of compas on GitHub.

\paragraph{\bf studentPerformance}
To preprocess studentPerformance, we dropped several categorical features: [`Mjob', `Fjob', `reason', `guardian', `school'] and encoded a binary target variable that labelled a data point as $1$ if the grade `G3' was greater than $11$. This led to a balanced dataset with chance accuracy of $0.5362$.

\paragraph{\bf oulad}
The features of the oulad dataset are already ordinal or binary. As the target, we used the feature `final\_result', which we binarised to be $1$ if a person passed and $0$ if not. We dropped missing values and thereby reached a chance accuracy of $0.582$.

\paragraph{\bf studentDropout and studentDropout\_binary} 
For this dataset we dropped several non-numerical or non-ordinal features: [`Application mode', `Course', `Nationality', `Father's occupation', `Mother's occupation', `Father's qualification', `Mother's qualification']. We further binarised `Marital status' and made `Previous qualification' ordinal. As target variable we used `Target', which we binarised to label all individuals as $1$ if they graduated and as $0$ if they dropped out. The dataset was balanced by subsampling to achieve a chance accuracy of $0.5$.

For the binary version, we further dropped the features [`Marital status', `Application order', `Previous qualification', `Gender', `Unemployment rate', `Previous qualification (grade)', `Educational special needs', `Tuition fees up to date', `GDP', `Inflation rate']. The remaining features were binned at the median feature value.

\paragraph{\bf breastCancer}
To download the data, we used \verb|load_breast_cancer()| from the \verb|sklearn.datasets| package for Python. The dataset was balanced by subsampling to achieve a chance accuracy of $0.5$.
 
\section{Additional information on the experiments in Section~\ref{sec:experiment_dataset_multiplicity}}\label{App:details_exp_data_mult}

\subsection{General procedure}
The purpose of the experiments in Section~\ref{sec:experiment_dataset_multiplicity} is to investigate the importance of dataset multiplicity for the expressiveness of the resulting Rashomon set.
Hence, for a fixed integration of parametric multiplicity we compare different ways of including dataset multiplicity.
For all different runs, parametric multiplicity is always implemented in the same way by using a grid of possible parameter combinations. For each model we randomly select a parameter combination without replacement from the grid and train the model with those hyperparameters. The experiments in the main paper investigate decision trees and use tree depth and splitting criterion as parameters. Section~\ref{App:additional_experiments_ACS} shows additional experiments with other model classes. Table~\ref{table:params_variation} gives an overview on the model classes and the parameters that were used to create the grid. While parametric multiplicity is fixed for all runs, we vary the integration of dataset multiplicity in levels from \textit{no dataset multiplicity} to \textit{full dataset multiplicity}. Specifically, we investigate bootstrapping, feature subsampling and adding feature noise.
\begin{table}[H]
\begin{tabular}{|l|l|l|l|l|l|}
\hline
\bfseries model & \bfseries parameter variations  \\ \hline\hline
decision tree (DT) & tree\_depth $\in \{1, \dots, 700\}$ \\ 
& splitting\_criterion $\in \{$gini, log\_loss, entropy$\}$ \\ \hline
XGBoost (XGB) & tree\_depth $\in \{1, \dots, 45\}$ \\
& n\_trees $\in \{1, \dots, 45\}$\\ \hline
random forest (RF) & tree\_depth $\in \{1, \dots, 45\}$ \\
& n\_trees $\in \{1, \dots, 45\}$ \\ \hline
multilayer perceptron (MLP)& n\_hidden\_layers $\in \{1, \dots, 26\}$\\
& size\_hidden\_layer $\in \{1, \dots, 26\}$\\
& activation $\in \{$relu, logistic, tanh$\}$ \\ \hline
logistic regression (logReg) & penalty $\in \{$l1, l2$\}$ \\
& $C \in \{0.01, \dots, 1.0, \dots, 100.0\}$\\ \hline
\end{tabular}
\caption{Parameter variations per model class. A grid of parameter values is created and for each grid point we train a model. The grid size is chosen in a way to end up with $\approx 2,000$ models.}
\label{table:params_variation}
\end{table}

\subsection{Determining the subset size for experiments with ACSEmployment}\label{App:accuracy_convergence}
\begin{figure}[H]
    \centering
    \includegraphics[width=0.7\textwidth]{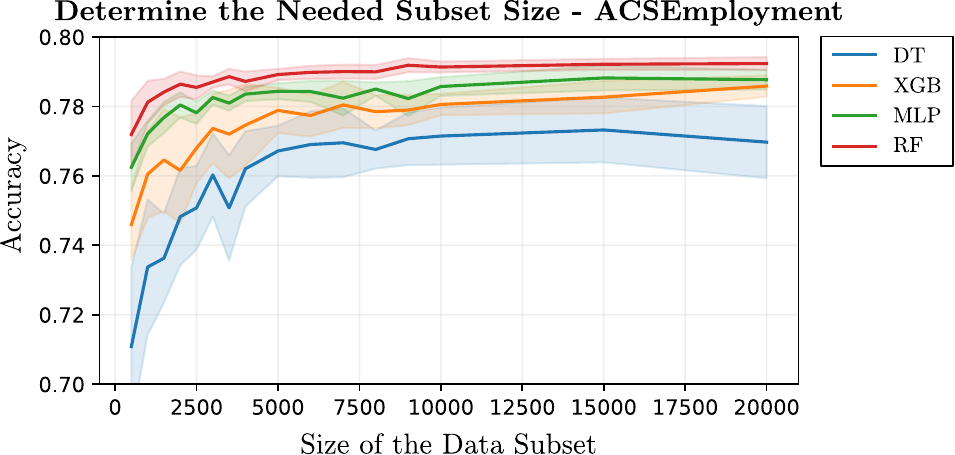}  
    \caption{
    The best accuracy is displayed depending on the size of the subset of the ACSEmployment data.
    }
    \label{fig:accuracies} 
\end{figure}

To determine the size of the subsets for our experiments with the ACSEmployment data, we looked into the accuracy of the models depending on the dataset size.
For each subset size we find the best accuracy via cross-validation.
We vary the same parameters that we use for model variations in the final experiments (see Section~\ref{sec:experiment_dataset_multiplicity}).
We can read off Figure~\ref{fig:accuracies} that the accuracy converges for a certain subset size.
Based on this, we decided $n=7,000$ for our experiments.

\subsection{Additional experiments on ACSEmployment}\label{App:additional_experiments_ACS}
\begin{figure}[H]
    \centering
    \includegraphics[width=\textwidth]{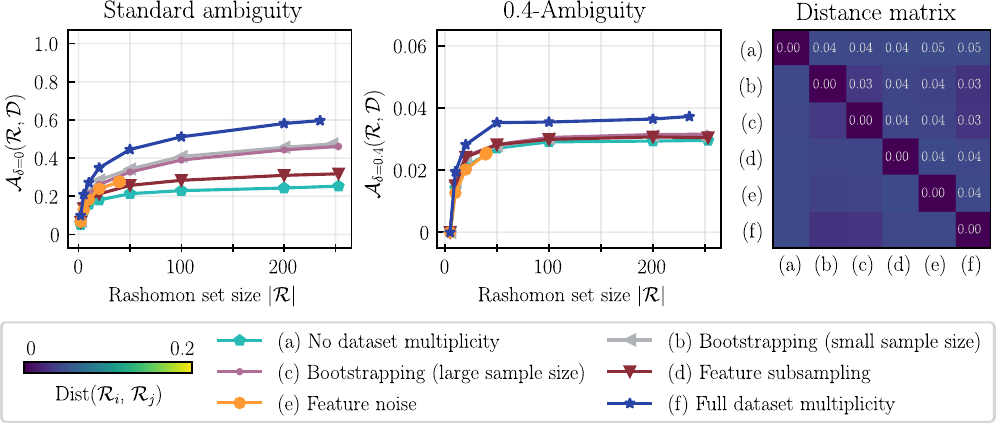}  
    \caption{
    \textbf{The importance of dataset multiplicity on ACSEmployment data for XGBoost.}
    For fixed parametric multiplicity, we compare different ways to include dataset multiplicity into the ad-hoc approach from (a) no dataset multiplicity to (f) full dataset multiplicity.
    Distances between approaches are depicted as heat map (right plot). 
    The baseline model has accuracy $acc(g_0) \approx 0.7926$ and we used $\varepsilon = 0.02$.
    }\label{fig:ACSEmployment_XGB}
\end{figure}

\begin{figure}[H]
    \centering
    \includegraphics[width=\textwidth]{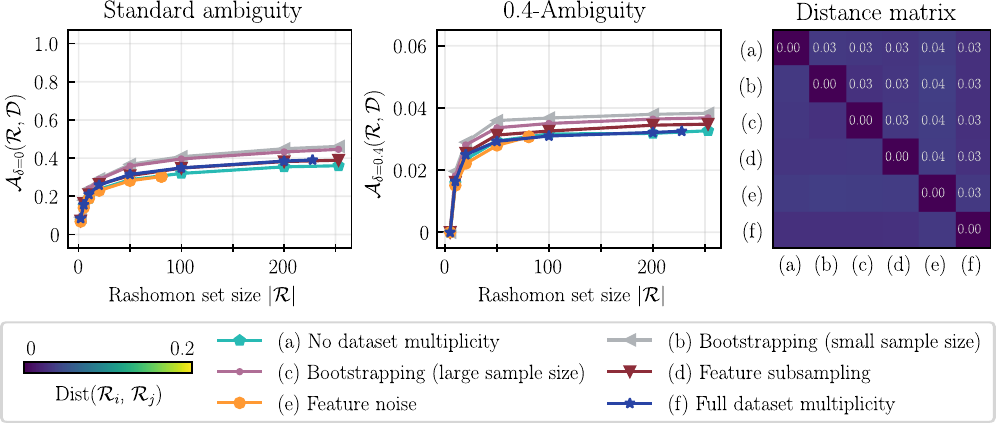}  
    \caption{
    \textbf{The importance of dataset multiplicity on ACSEmployment data for MLPs.}
    For fixed parametric multiplicity, we compare different ways to include dataset multiplicity into the ad-hoc approach from (a) no dataset multiplicity to (f) full dataset multiplicity.
    Distances between approaches are depicted as heat map (right plot). 
    The baseline model has accuracy $acc(g_0) \approx 0.7900$ and we used $\varepsilon = 0.02$.
    }\label{fig:ACSEmployment_MLP}
\end{figure}

To verify our finding of the importance of dataset multiplicity beyond decision trees, we present results of further model classes.
We conduct the same experiment from above (Section~\ref{sec:experiment_dataset_multiplicity}) on the ACSEmployment data for XGBoost (Figure~\ref{fig:ACSEmployment_XGB}) and for MLPs (Figure~\ref{fig:ACSEmployment_MLP}).
Similar to the above findings, including dataset multiplicity improves the ability of the resulting Rashomon set to detect conflicting and highly conflicting data points.

\subsection{Additional experiments on various other datasets}\label{App:additional_experiments_further_datasets}
\begin{figure}[H]
    \centering
    \includegraphics[width=\textwidth]{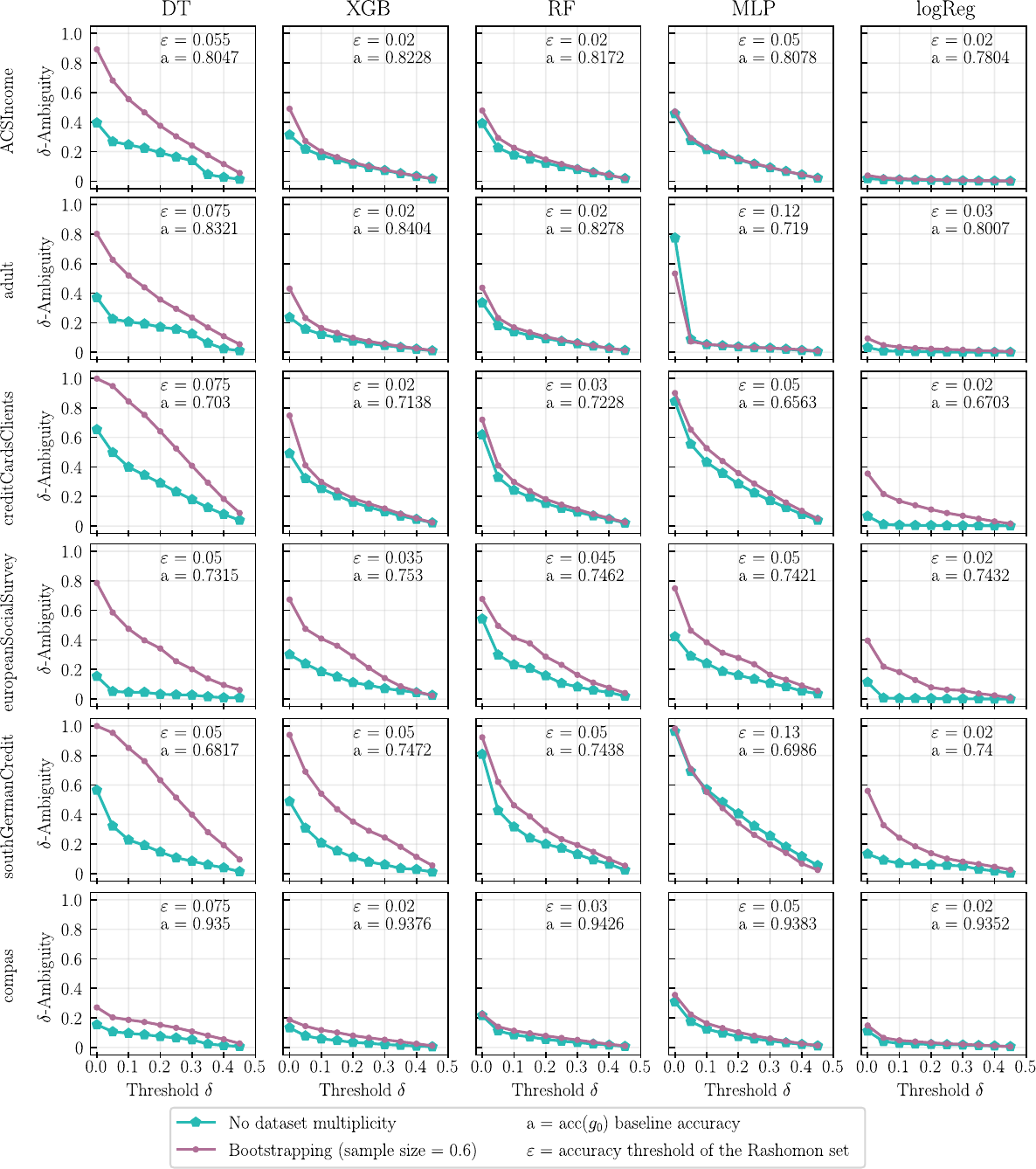}  
    \caption{\textbf{The importance of dataset multiplicity across various datasets and model classes:}
    For each dataset (rows) we use five different model classes (columns) to calculate Rashomon sets of size $200$ via the ad-hoc approach.
    $\delta$-ambiguity curves are plotted for the \textit{no dataset multiplicity} condition and \textit{bootstrapping with sample size $0.6$}.
    A higher curve represents a better ability to detect conflicting and highly conflicting data points. 
    In most cases including dataset multiplicity in form of bootstrapping leads to higher $\delta$-ambiguity curves. 
    }\label{fig:ambiguity-summaryA}
\end{figure}

\begin{figure}[H]
    \centering
    \includegraphics[width=\textwidth]{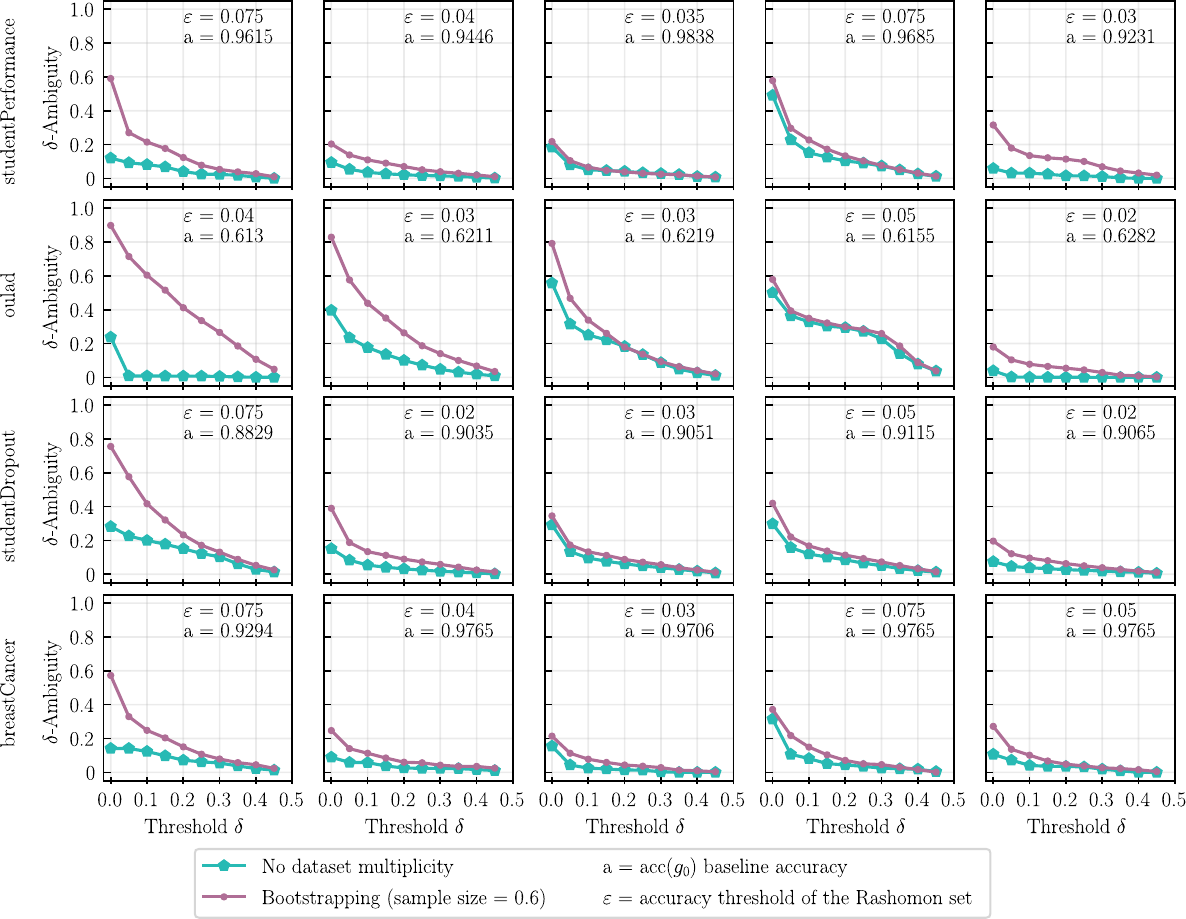}  
    \caption{\textbf{The importance of dataset multiplicity across various datasets and model classes:}
    For each dataset (rows) we use five different model classes (columns) to calculate Rashomon sets of size $200$ via the ad-hoc approach.
    $\delta$-ambiguity curves are plotted for the \textit{no dataset multiplicity} condition and \textit{bootstrapping with sample size $0.6$}.
    A higher curve represents a better ability to detect conflicting and highly conflicting data points. 
    In most cases including dataset multiplicity in form of bootstrapping leads to higher $\delta$-ambiguity curves.
    }\label{fig:ambiguity-summaryB}
\end{figure}

We want to see whether the importance of dataset multiplicity generalises beyond the specific choice of \mbox{ACSEmployment} as dataset and therefore run further experiments for several model classes and datasets.
As model classes, we use XGBoost, random forests, decision trees, MLPs, and logistic regression. 
See Table~\ref{table:params_variation} for a list of model parameters that we vary to induce parametric multiplicity.
An overview of the datasets can be found in Table~\ref{table:datasets}.
Since in the above experiments, bootstrapping showed promising ability to induce dataset multiplicity into the ad-hoc approach, we concentrate on the two approaches \textit{no dataset multiplicity} and \textit{bootstrapping with sample size $0.6$}.
The achieved $\delta$-ambiguity curves are displayed in Figures~\ref{fig:ambiguity-summaryA} and \ref{fig:ambiguity-summaryB}.
For each pair of model and dataset the value for $\varepsilon$ is chosen in a way that at least $200$ out of $2000$ trained models meet the accuracy threshold of the Rashomon set.
If the Rashomon set is larger than $200$ models, then for comparison, the $\delta$-ambiguity values are calculated for $200$ models randomly chosen out of the Rashomon set.
In nearly all cases, the $\delta$-ambiguity curves of the \textit{bootstrapping} approach are above those of \textit{no dataset multiplicity}. 
This confirms our finding from above that the inclusion of dataset multiplicity helps in detecting conflicting and highly conflicting data points.

\section{Additional information on the experiment in Section~\ref{sec:experiment_treefarms}}\label{App:details_exp_treefarms}
The TreeFarms algorithm \citep{Xin_2022_ExploringTheWholeRashomonSet} enforces sparsity of the decision trees with a penalty on the number of leaves $H(g)$ in their objective function,
\begin{align*}
    Obj(g) = acc(g) - \lambda H(g).
\end{align*}
The corresponding regularisation parameter $\lambda$ determines how much weight the penalty gets, with a higher value leading to more sparsity.
For our experiments we choose $\lambda \in [0, 1]$ separately per dataset such that the found Rashomon sets are of roughly the same size for all three datasets considered (details in Table~\ref{table:TreeFarms_parameters}). \\
To ensure comparability, we use the same objective function for the ad-hoc approach.
Hence, a model $g$ is included into the Rashomon set if $Obj(g) \geq Obj(g_0) - \varepsilon$, which slightly changes the Rashomon set compared to Definition~\ref{def:rashomon_set}.
As baseline model $g_0$, we select the model with best objective value $Obj(g_0)$ found by TreeFarms (details in Table~\ref{table:TreeFarms_parameters}).
The Rashomon set threshold is set to $\varepsilon = 0.05$ for all three datasets.
\\
We consider parametric multiplicity by varying depth, splitting criteria, and random seed of the decision trees.
To incorporate dataset multiplicity we bootstrap data subsamples of size $0.2 \cdot \lvert \mathcal{D}_{\textrm{train}} \rvert$.

\begin{table}[H]
\begin{tabular}{|l|l|l|l|l|l|}
\hline
\bfseries dataset & \bfseries $\lambda$ & \bfseries $\epsilon $  & \bfseries $Obj(g_0)$  & \bfseries $acc(g_0)$ & \bfseries $H(g_0)$\\ \hline\hline
europeanSocialSurvey & $0.015$ & $0.05$ & $0.6832$ & $0.7432$ & $4$ \\ \hline
studentDropout\_binary & $0.013$ & $0.05$ & $0.7749$ & $0.8269$ & $4$ \\ \hline
compas\_binary & $0.01$  & $0.05$ & $0.6236$ & $0.6536$ & $3$ \\ \hline
\end{tabular}
\caption{Parameters and resulting best model $g_0$ for each of the datasets used in the experiments of Section~\ref{sec:experiment_treefarms}. The regularisation parameter $\lambda$ influences how much weight the number of leaves $H(g)$ and the accuracy $acc(g)$ get in the objective function $Obj(g)$. The objective result for the baseline model $Obj(g_0)$ together with $\varepsilon$ determines the threshold of the Rashomon set.}
\label{table:TreeFarms_parameters}
\end{table}

\section{Code availability}
The code used for the experiments in Sections~\ref{sec:experiment_dataset_multiplicity}, \ref{sec:experiment_treefarms} and \ref{App:additional_experiments_further_datasets} can be found at \url{https://github.com/KaroFr/PM_AIA}.

\end{document}